\pdfoutput=1

\documentclass[11pt]{article}

\usepackage{EACL2023}

\usepackage{times}
\usepackage{latexsym}

\usepackage[T1]{fontenc}
\usepackage[utf8]{inputenc}

\usepackage{microtype}

\usepackage{inconsolata}

\usepackage{amssymb}% http://ctan.org/pkg/amssymb
\usepackage{pifont}% http://ctan.org/pkg/pifont

\usepackage{pbox}
\usepackage{hyperref}
\usepackage{url}
\usepackage{comment}
\usepackage[flushleft]{threeparttable}
\usepackage{graphicx}
\usepackage{amsmath}
\usepackage{tabularx}
\usepackage{multirow, makecell}
\usepackage{algorithm}
\usepackage{xcolor}
\usepackage{booktabs}
\usepackage{paralist}
\usepackage{caption}
\usepackage{subcaption}
\usepackage[noend]{algpseudocode}
\usepackage{color, colortbl}

\usepackage{amsthm}
\theoremstyle{definition}

\usepackage[noabbrev]{cleveref}
\crefname{section}{§}{§§}
\Crefname{section}{§}{§§}

\definecolor{MyColor}{RGB}{50, 100, 250}
\definecolor{Orange}{RGB}{244, 101, 66}
\definecolor{Red}{RGB}{255, 0, 0}
\definecolor{Green}{RGB}{0, 255, 0}
\definecolor{Blue}{RGB}{0, 0, 255}

\newcommand{\mytilde}{\raise.17ex\hbox{$\scriptstyle\mathtt{\sim}$}}

\usepackage{arydshln}
\usepackage{listings}
\usepackage{adjustbox}
\usepackage[most]{tcolorbox}
\usepackage{xspace}
\newcommand{\sng}{$\mathcal{S} \& \mathcal{G}$\xspace}

\newcommand{\eg}{\emph{e.g.,}\xspace}
\newcommand{\ie}{\emph{i.e.,}\xspace}

\newcommand{\wrt}{\emph{w.r.t.}\xspace}

\definecolor{dark-gray}{gray}{0.85}
\definecolor{light-gray}{gray}{0.95}
\definecolor{mygreen}{rgb}{0,0.4,0}
\definecolor{mygray}{rgb}{0.5,0.5,0.5}
\definecolor{mymauve}{rgb}{0.58,0,0.82}
\definecolor{myred}{rgb}{0.82, 0.1, 0.26}

\lstdefinestyle{CustomPy}{
    escapeinside={(*@}{@*)},
    belowcaptionskip=1\baselineskip,
    xleftmargin=1pt,
    xrightmargin=1pt,
    language=Python,
    numbersep=5pt,
    tabsize=4,
    showstringspaces=false,
    basicstyle=\small\ttfamily, % basic font setting %Previously it was sffamily, changed to ttfamily
    keywordstyle=\bf\color{mygreen},
    commentstyle=\color{purple},
    stringstyle=\color{red},
    identifierstyle=\color{black},
    numberstyle=\tiny\color{mygray},
    emph={int,char,double,float,unsigned,void,bool,boolean},
    emphstyle={\bf\color{myred}},
    emph=[2]{and, in,},
    emphstyle=[2]{\bf\color{violet}},
    emph=[3]{sortedCount, sorted_count},
    emphstyle=[3]{\bf\color{blue}},
    numbers=left,
    stepnumber=1,
    breaklines=true,
    backgroundcolor=\color{white},
    literate={\ \ }{{\ }}1,
}

\lstdefinestyle{CustomJava}{
    belowcaptionskip=1\baselineskip,
    xleftmargin=1pt,
    xrightmargin=3pt,
    language=Java,
    numbersep=5pt,
    tabsize=4,
    showstringspaces=false,
    basicstyle=\small\ttfamily, % basic font setting %Previously it was sffamily
    keywordstyle=\bf\color{mygreen},
    commentstyle=\color{purple},
    stringstyle=\color{red},
    identifierstyle=\color{black},
    numberstyle=\tiny\color{mygray},
    stringstyle=\color{mymauve},
    emph={int,char,double,float,unsigned,void,bool,boolean},
    emphstyle={\bf\color{myred}},
    emph=[2]{and, in,},
    emphstyle=[2]{\bf\color{violet}},
    emph=[3]{sortedCount, sorted_count},
    emphstyle=[3]{\bf\color{blue}},
    numbers=left,
    stepnumber=1,
    breaklines=true,
    backgroundcolor=\color{white},
    literate={\ \ }{{\ }}1,
}

\makeatletter
\let\old@lstKV@SwitchCases\lstKV@SwitchCases
\def\lstKV@SwitchCases#1#2#3{}
\makeatother
\usepackage{lstlinebgrd}
\makeatletter
\let\lstKV@SwitchCases\old@lstKV@SwitchCases

\lst@Key{numbers}{none}{%
    \def\lst@PlaceNumber{\lst@linebgrd}%
    \lstKV@SwitchCases{#1}%
    {none:\\%
     left:\def\lst@PlaceNumber{\llap{\normalfont
                \lst@numberstyle{\thelstnumber}\kern\lst@numbersep}\lst@linebgrd}\\%
     right:\def\lst@PlaceNumber{\rlap{\normalfont
                \kern\linewidth \kern\lst@numbersep
                \lst@numberstyle{\thelstnumber}}\lst@linebgrd}%
    }{\PackageError{Listings}{Numbers #1 unknown}\@ehc}}
\makeatother

\newcount\myloopcounter

\newcommand{\repeatit}[2][10]{%
  \myloopcounter0% initialize the loop counter
  \loop\ifnum\myloopcounter < #1 % Test if the loop counter is < #1
  #2%
  \advance\myloopcounter by 1 % 
  \repeat % start again
}

\usepackage{titlesec}
\titlespacing{\paragraph}{%
  0pt}{%              left margin
  0.2\baselineskip}{% space before (vertical)
  1em}%

\title{Summarize and Generate to Back-translate: \\ Unsupervised Translation of Programming Languages}

\author{
Wasi Uddin Ahmad$^\dagger$, Saikat Chakraborty$^*$, Baishakhi Ray$^\ddagger$, Kai-Wei Chang$^\dagger$ \\
$^\dagger$University of California, Los Angeles, $^*$Microsoft Research, $^\ddagger$Columbia University \\ [2pt]
$^\dagger$\texttt{\{wasiahmad,kwchang\}@ucla.edu}  $^*$\texttt{saikatc@microsoft.com} $^\ddagger$\texttt{rayb@cs.columbia.edu}
}

\begin{document}
\maketitle

\setlength{\abovedisplayskip}{5pt}
\setlength{\belowdisplayskip}{5pt}

\begin{abstract}
Back-translation is widely known for its effectiveness in neural machine translation when there is little to no parallel data. In this approach, a source-to-target model is coupled with a target-to-source model trained in parallel. The target-to-source model generates noisy sources, while the source-to-target model is trained to reconstruct the targets and vice versa. Recent developments of multilingual pre-trained sequence-to-sequence models for programming languages have been very effective for a broad spectrum of downstream software engineering tasks. Hence, training them to build programming language translation systems via back-translation is compelling. However, these models cannot be further trained via back-translation since they learn to output sequences in the same language as the inputs during pre-training. As an alternative, we propose performing back-translation via code summarization and generation. In code summarization, a model learns to generate natural language (NL) summaries given code snippets. In code generation, the model learns to do the opposite. Therefore, target-to-source generation in back-translation can be viewed as a \emph{target-to-NL-to-source} generation. We show that our proposed approach performs competitively with state-of-the-art methods.
We have made the code publicly available.\footnote{\url{https://github.com/wasiahmad/SumGenToBT}}

\end{abstract}

\section{Introduction}
\label{sec:intro}

The choice of programming language (PL) in software development depends on the requirement of the software and the available features of a particular PL. In modern API-driven software development, language choice often depends on the availability of libraries and APIs.
The advent of newer and richer programming languages often requires legacy software to be translated into modernized PLs. 
In theory, modern programming languages' ``Turing Completeness'' allows rule-based translation of programs from one PL to another. 
The rule-based translation may require an extensive number of custom-written transformation rules and could end up producing very unreadable source code. In addition, such a translation could entail translating the entire library, even if a library implementing similar functionality is available in the target PL.  

\begin{figure}[t]
% \rule{\linewidth}{0.6pt}
\centering
\begin{adjustbox}{valign=t,minipage=0.47\textwidth}
\begin{center}
    \underline{Input in Java}
\end{center}
\vspace{-5pt}
\begin{tabular}{@{}l}
\lstset{escapechar=~,style=CustomJava}
\begin{lstlisting}
public static boolean generator(PsiBuilder b, int l){
    boolean r = false;
    if(!recursion_guard(b, l)) return r;
    r = generator_0(b, l + 1);
    if(!r) return generator_1(b, l + 1);
}
\end{lstlisting}
\end{tabular}
\end{adjustbox}
\\
\vspace{5pt}
\begin{adjustbox}{valign=t,minipage=0.47\textwidth}
%  (w/o fine-tuning)
\begin{center}
    \underline{\normalsize (1) Java to Python Generation}
\end{center}
\vspace{-5pt}
\begin{tabular}{@{}l}
\lstset{escapechar=~,style=CustomJava}
\begin{lstlisting}[ 
    linebackgroundcolor={%
    \ifnum\value{lstnumber}=4
        \color{red!10}
    \fi
    }
]
public static boolean generator(PsiBuilder b, int l){
    boolean r = false;
    if(!recursion_guard(b, l)) return r;
    if(!generator_0(b, l)) return r;
    r = generator_0(b, l + 1);
    if(!r) return generator_1(b, l + 1);
}
\end{lstlisting}
\end{tabular}
\end{adjustbox}
\\
% \vspace{5pt}
% \begin{center}
%     \normalsize{\color{blue} Summarize and Generate to Back-translate}
% \end{center}
% \vspace{10pt}
% \rule{\linewidth}{0.6pt}
\begin{center}
    \underline{\normalsize (2) Java Code to NL Summary}
\end{center} 
\vspace{-7pt}
\begin{flushleft}
guard is used to determine if a generator is already defined .
\end{flushleft}
% \vspace{5pt}
\begin{adjustbox}{valign=t,minipage=0.47\textwidth}
\begin{center}
    \underline{\normalsize (3) NL Summary to Python Code}
\end{center} 
\vspace{-5pt}
\begin{tabular}{@{}l}
\lstset{escapechar=@,style=CustomPy}
\begin{lstlisting}
def is_generator(self, name):
    if name in self._generators:
        return True
    if name in self._generators[name]:
        return True
    return False
\end{lstlisting}
\end{tabular}
\end{adjustbox}
% \vspace{-1mm}
\caption{
Although PLBART is asked to generate in Python given input in Java (1), it generates in Java (due to its pre-training objective). In contrast, PLBART, fine-tuned on code summarization and generation, generates ``noisy'' translations (as in (2, 3)).
}
% \vspace{-2mm}
\label{fig:qual_ex1}
\end{figure}

\begin{figure*}[t]
	\centering
	\subfloat[PLBART \label{sfig:plbart}]{%
		\includegraphics[width=0.49\textwidth]{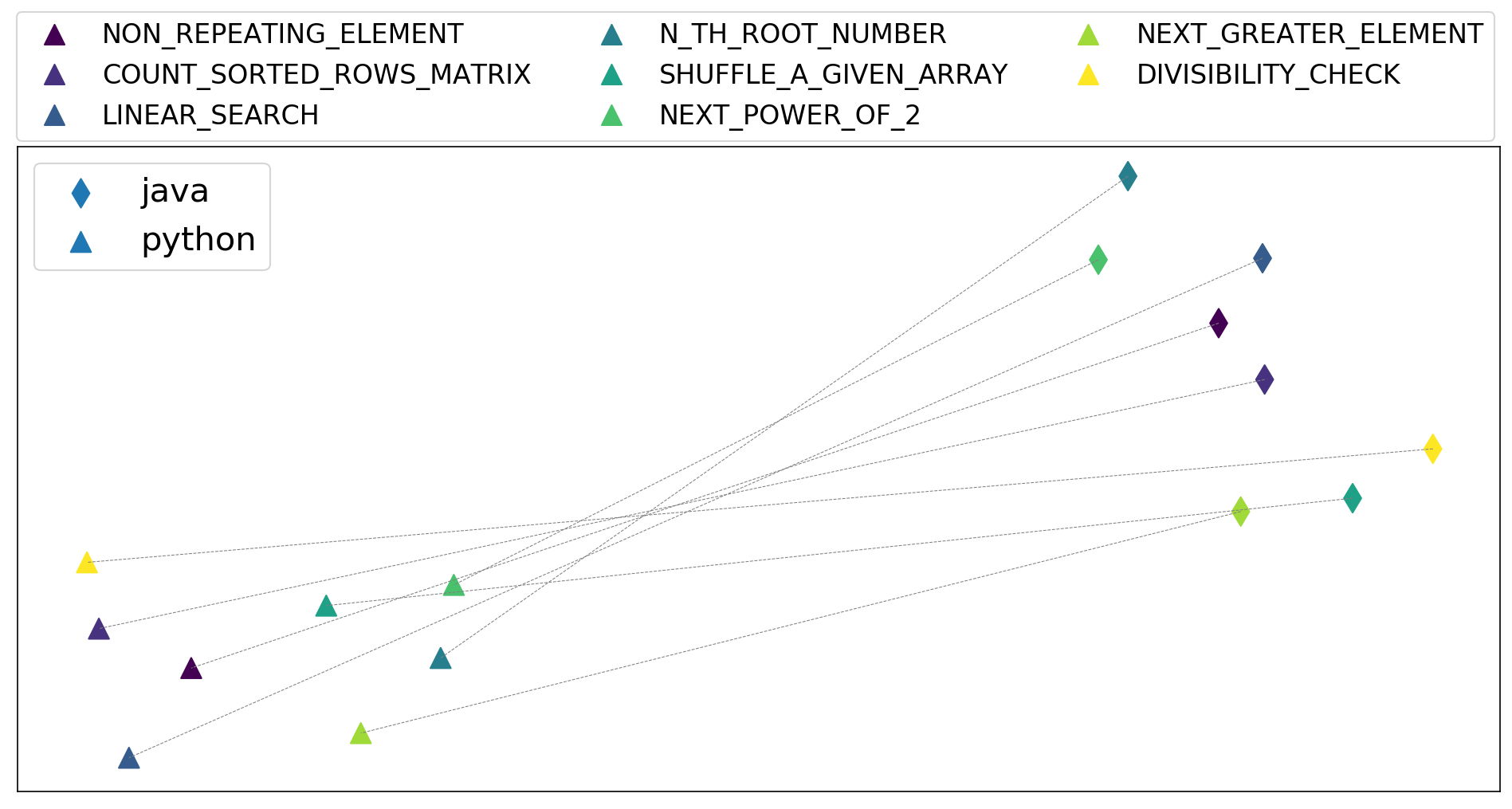}
	}
	\subfloat[PLBART + \sng \label{sfig:plbart_msg}]{%
		\includegraphics[width=0.49\textwidth]{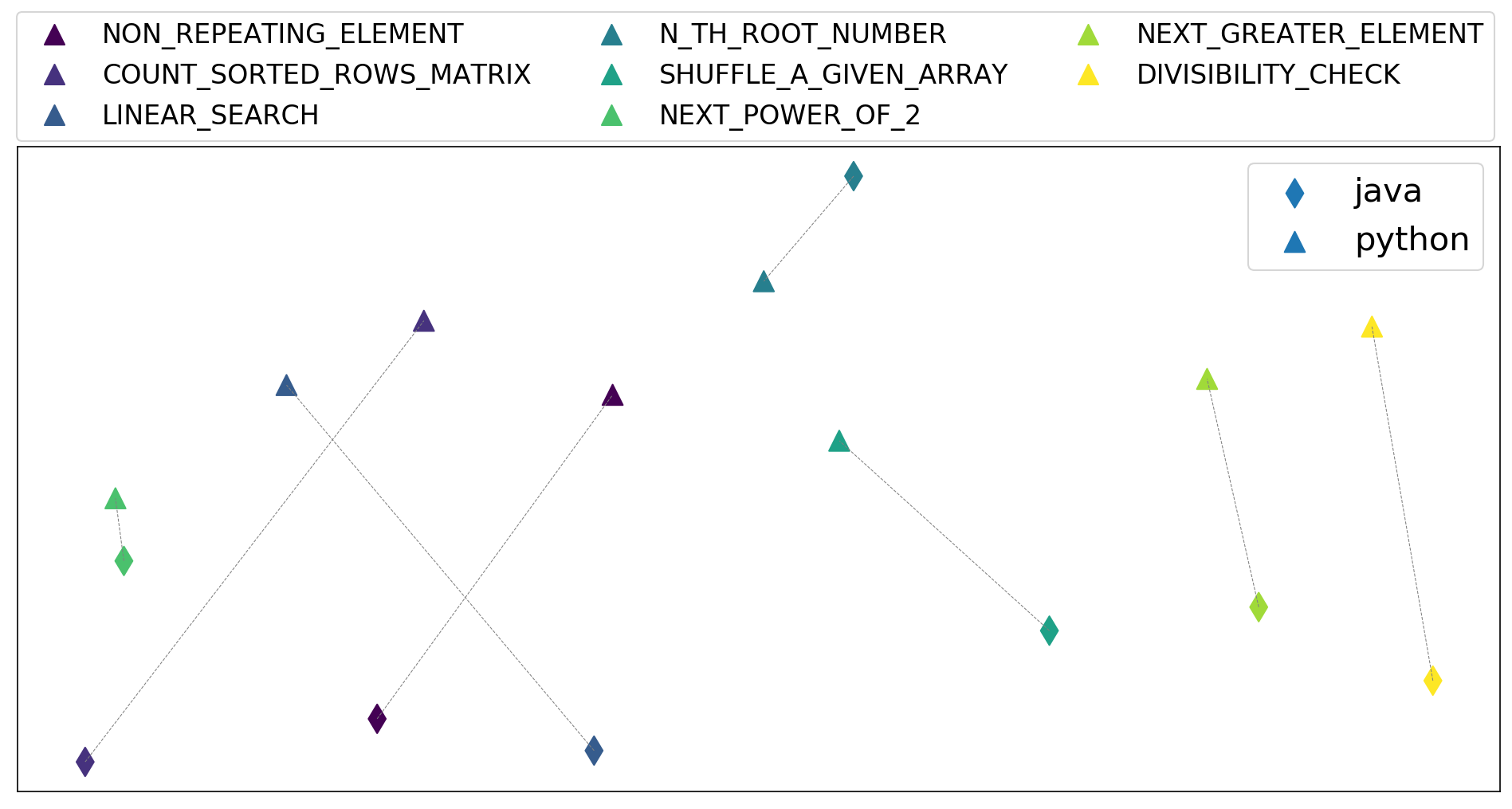}
	}
% 	\vspace{-2mm}
	\caption{T-SNE plot of function embeddings of Java and Python functions. \Cref{sfig:plbart} shows the embedding generated by the PLBART model. 
	\Cref{sfig:plbart_msg} are the generated embedding when the PLBART is finetuned to jointly summarize code to NL and generate code from NL (PLBART + \sng). 
% 	The average distance between parallel java and python code is 343.01 as generated by PGLM and 230.74 as generated by PGLM + S\&G.
    While PLBART clusters programs from each PLs, parallel programs in different PLs are brought closer to each other by PLBART + \sng.
	}
	\label{fig:tsne}
% 	\vspace{-2mm}
\end{figure*}

Aligning libraries and APIs across different PLs is a non-trivial task. Recent progress in Neural Machine Translation (NMT) \cite{bahdanau2014neural, vaswani2017attention} leveraging pre-trained models~\cite{feng-etal-2020-codebert, guo2020graphcodebert, roziere2021dobf, ding2021contrastive,ahmad-etal-2021-unified, wang-etal-2021-codet5} could be a possible way to learn the alignment between PLs and translate source code across languages.
% demonstrated program understanding~\cite{} and generation capacity~\cite{}. 
% However, the NMT models should exhibit a profound understanding of the source language and a reasonably well generation capacity to translate programs across languages. There are several recently proposed pretrained models that demonstrated program understanding~\cite{feng-etal-2020-codebert, guo2020graphcodebert, roziere2021dobf, ding2021contrastive} and generation capacity~\cite{ahmad-etal-2021-unified, wang-etal-2021-codet5}.

A significant challenge in supervised learning for NMT is the need for large-scale parallel corpora. 
For instance, if we are planning to train a translator for {\tt Java} to {\tt Python} translation, we need a considerable number of the same program (\emph{i.e.}, exhibiting the same semantic behavior) in both languages. Availability of such parallel datasets is a vital challenge in programming language translation~\cite{chen2018tree}. Back-Translation (BT)~\cite{edunov-etal-2018-understanding, lachaux2020unsupervised} is a clever way to learn alignments across different languages.
While BT demonstrates success in NMT, those require either (i.) small (perhaps noisy) parallel datasets or (ii.) a model with some capacity of cross-lingual generation - to kickstart the BT-based learning process.

In this work, we investigate the suitability of the multilingual Pre-trained Sequence-to-Sequence Model (PSM) 
% (\emph{e.g.}, PLBART~\cite{ahmad-etal-2021-unified}) 
for unsupervised programming language translation via BT. In particular, we assume a use-case scenario where {\em no} parallel data is available. Without much of a surprise, we empirically found that, while these PSMs are good at generating code in each language, they exhibit very little to no knowledge about cross-lingual generation since such PSMs are typically trained to reconstruct code sequences from noisy inputs.
For example, when we provide the input code in \Cref{fig:qual_ex1} to PLBART~\cite{ahmad-etal-2021-unified} and ask to generate Python code without training, it generates a slight variation of the input Java code, showing its lack of knowledge about cross-lingual generation.

To endow such PSMs with knowledge about cross-lingual generation, we propose using a third language ({\em i.e.}, English).
Since a large quantity of monolingual code corpora comes with documentation, which supposedly describes what the source code is doing, we train a Summarize-and-Generate (\sng) model that can generate pseudo-parallel code sequences.
\Cref{fig:qual_ex1} shows PLBART's behavior when it is further trained via \sng. First, given the Java code, it generates an NL summary (\cref{fig:qual_ex1}-2) and subsequently generates Python Code (\cref{fig:qual_ex1}-3). 
We empirically show that, even if such \sng model generates noisy parallel sequences, it allows us to employ PSMs in the BT-based training to learn programming language translation.
% it significantly improves the translation performance of PSMs using BT.

In summary, we present a Summarize-and-Generate (\sng) based approach to enable unsupervised program translation training of PLBART via Back-Translation (BT).
% of pretrained sequence-to-sequence source code models.
% (\eg PLBART~\cite{ahmad-etal-2021-unified}). 
Experiment results show that our proposed approach makes PLBART
% pre-trained sequence-to-sequence model 
trainable via BT and performs competitively with state-of-the-art program translation models.

% \footnote{We have made our code publicly available at \url{https://github.com/hidden/hidden}.}
% https://github.com/wasiahmad/SumGenToBT
% \url{https://github.com/hidden/hidden}.}
% In-depth analysis shows that with \sng training, PLBART performs comparably or better than state-of-the-art unsupervised program translator.

\section{Motivation}
\label{sec:motivation}

% \begin{figure*}[t]
%     \centering
%     \includegraphics[width=0.85\textwidth]{images/overview.pdf}
%     \caption{
%         Overview of our proposed back-translation framework to train PLBART. In the first $m$ steps (out of total $N$ training steps), we use a multilingual code summarization and generation model ($\mathcal{S}, \mathcal{G}$) to perform back-translation. In the remaining steps ($N-m$), PLBART is trained via standard back-translation method.
%     }
%     \label{fig:overview}
% \end{figure*}

\begin{figure*}[t]
	\centering
    \begin{subfigure}{0.75\textwidth}
    \centering
    \includegraphics[width=\textwidth]{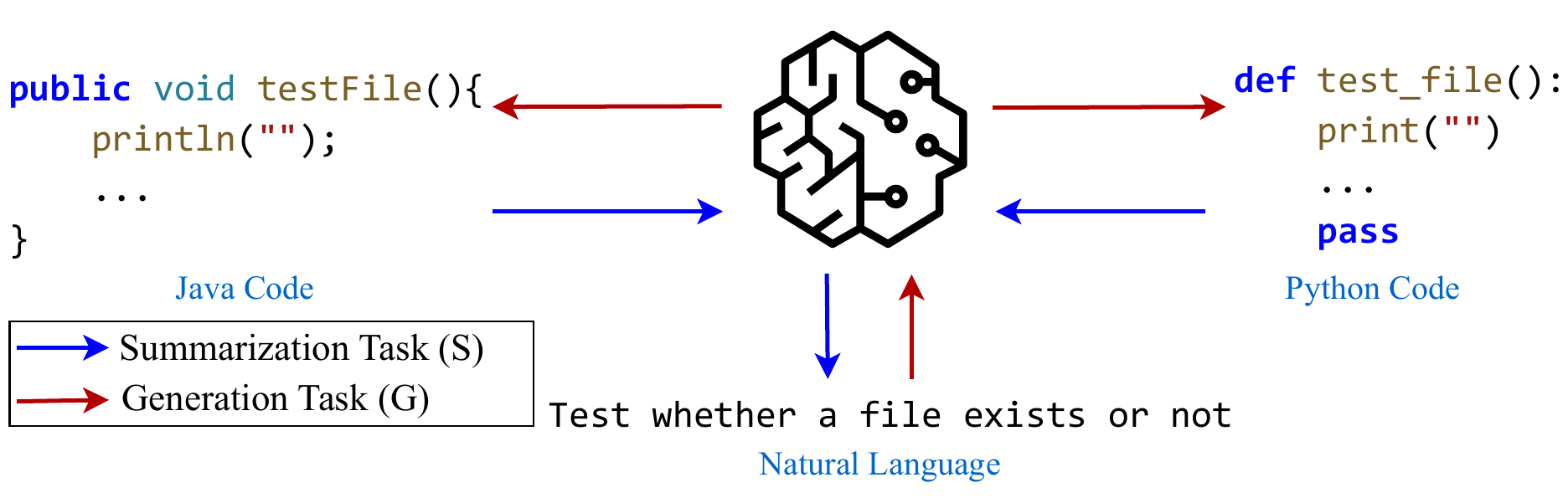}
    \caption{Step1: Supervised training of PLBART on Code Summarization and Generation (\sng).}
    \label{sfig:overview-s-n-g}
    \end{subfigure}
    
    \vspace{10pt}
    
    \begin{subfigure}{0.90\textwidth}
    \centering
    \includegraphics[width=\textwidth]{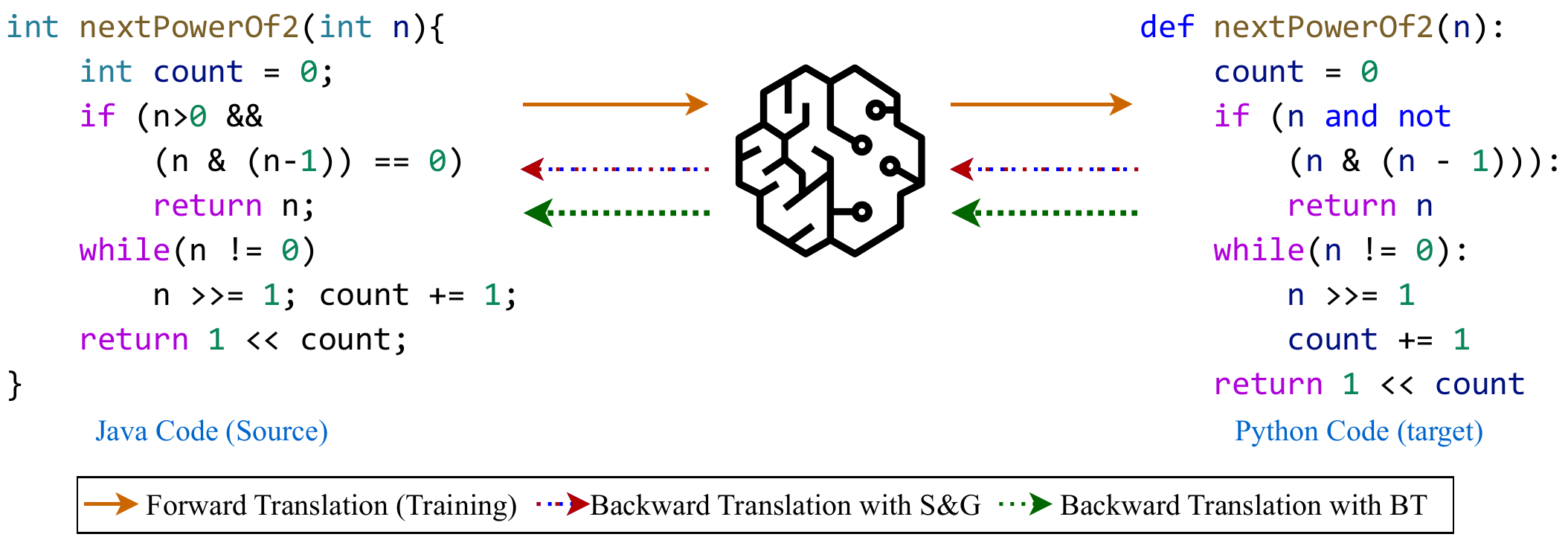}
    \vspace{-2mm}
    \caption{Step2: Unsupervised training of PLBART via Backtranslation (BT). In the first $m$ training steps (out of total $N$ steps), PLBART generates natural language (NL) summary of the code in target language (Python in this example) and generates the code in source language (Java in this example) from the NL summary. In the remaining $N-m$ steps, PLBART directly generates the code in source language from the code in target language.}
    \label{sfig:overview-bt}
    \end{subfigure}
	\caption{Overview of our proposed framework to train PLBART in two sequential steps.
	}
	\label{fig:overview}
\end{figure*}

Recent years saw several Pre-trained Sequence-to-Sequence models (PSM) \cite{ahmad-etal-2021-unified, wang-etal-2021-codet5}. These models are pre-trained on hundreds of Gigabytes of source code. Thus, in this work, we are motivated to investigate their adoption in learning program translation via back-translation.
% in the back-translation approach. In particular, we investigate whether we can directly train these pre-trained models for program translation via back-translation. 
To understand such feasibility, we investigate the program representations generated by the PSM. 
% As a case study, we chose PLBART \cite{ahmad-etal-2021-unified}. 
% Motivated by Tatoeba \cite{artetxe-schwenk-2019-massively} for understanding representation learning behavior of multilingual models, 
% we probe into the function embeddings given by PSMs.
As a case study, we chose PLBART \cite{ahmad-etal-2021-unified} and evaluated its multilingual embeddings as suggested in \citet{artetxe-schwenk-2019-massively}.
% Motivated by Tatoeba \cite{artetxe-schwenk-2019-massively} for understanding representation learning behavior of multilingual models, we probe into the function embeddings given by PSMs.
We find the parallel Java function for each of the 948 Python functions using the parallel dataset proposed in \citet{lachaux2020unsupervised}.
We find the nearest neighbor using cosine similarity between function embeddings and calculate the error rate.
Unsurprisingly, PLBART performs poorly in function retrieval with an 87.5\% error rate.
% Without much of a surprise, we find that in {\em only 12.5\%} cases (top-1 retrieval accuracy),
% \footnote{top-1 retrieval accuracy.} 
% the closest function is the corresponding Java function.

In comparison, we fine-tune PLBART jointly on code summarization and generation in Java and Python. 
Repeating the experiment of function retrieval, we find that fine-tuned PLBART's error rate drops to 23.7\%.
To visually illustrate the embeddings produced by PLBART and its fine-tuned variant, we provide a T-SNE plot of 8 sample functions' embedding in Figure \ref{fig:tsne}.
We see the functions that belong to the same language are clustered together while the same functions in two different languages are far apart from each other (see \Cref{sfig:plbart}).

% , which shows programs in the same language are clustered together. The embeddings of the same function in different languages are far apart. 

In contrast, the fine-tuned PLBART breaks up the intra-language clusters and brings functions in different languages close to each other in the embedding space (see \Cref{sfig:plbart_msg}).
These results motivate us to initialize the translation models with fine-tuned PLBART on code summarization and generation for back-translation as it learned some alignment across programming languages.

% These results suggest that fine-tuned PLBART on code summarization and generation learns some alignment across programming languages, which we consider as motivation to kickstart the back-translation by initializing the translation models with fine-tuned PLBART.

% We further train the PSM to simultaneously Summarize-and-Generate (\sng) and repeat our previous experiment on understanding the program embedding. After \sng training, the PSM generated embedding brings  {\em 76.3\%} programs from different languages to be the closest to each other. As shown in \cref{sfig:plbart_msg}, \sng training breaks up the intra-language clusters learned by PSM and brings programs in different languages close to each other. We also observe a similar trend when finding the closest python function given the java function embeddings. Nevertheless, since \sng training endows PSM with some initial alignment across languages, we are motivated to use \sng trained PSM to kickstart the back-translation. 

% \begin{figure*}[t]
%     \centering
%     \includegraphics[width=0.85\textwidth]{images/overview.pdf}
%     % \vspace{-2mm}
%     \caption{
%     Overview of our proposed back-translation framework to train PLBART. In the first $m$ steps (out of total $N$ training steps), we use a multilingual code summarization and generation model ($\mathcal{S}, \mathcal{G}$) to perform back-translation. In the remaining steps ($N-m$), PLBART is trained via standard back-translation method.
%     }
%     % \vspace{-2mm}
%     \label{fig:overview}
% \end{figure*}
%  Figure has been moved to mid of the motivation section. 

\section{Approach}
\label{sec:method}

% \notewa{One of our NAACL reviewers said, ``this section reads in a round-about way, starting with defining back-translation, and then code summarization and generation, and then back-translation again. It will read better if they start with detailing their code summarization and generation and then followed by how they use this model fine-tuned for code summarization and generation to do back-translation; instead of repeating the back-translation description twice. Another aspect that can be improved is in terms of notation. In Section 3, all the equations have consistently referred to source code as x and target code as y. However, the algorithm section (Algorithm 1) uses different notations $\_s$ or $\_t$ (for source and target, consecutively -- with $x$ and $y$ used interchangeably). This has made the reading of the Algorithm difficult and inconsistent with the in-text equations (eq. 1 and eq. 3). If the authors can make the notation in Algorithm 1 more consistent with the equations, that will improve readability greatly.''}

Sequence-to-sequence models, such as PLBART \cite{ahmad-etal-2021-unified}, CodeT5 \cite{wang-etal-2021-codet5}, SPT-Code~\cite{niu2022spt} map source code sequences into a shared multilingual space by pre-training on multiple programming languages jointly using unlabeled data ({\em e.g.}, source code from Github).
% due to jointly pre-training in many languages
The pre-training objective of these models is either denoising autoencoding (DAE) or fill-in-the-blank, where the models reconstruct the original code snippet or predict the missing code tokens given a corrupted code snippet. Although pre-trained jointly on many languages, these models only learn to generate in the same language as input.
% \footnote{For example, given a code snippet in Java language as input and setting ``<python>'' as the decoder prefix token (language ID), PLBART still generates code in Java.}
As a result, these models are not trainable via back-translation (BT) to learn programming language translation in an unsupervised fashion.
As an alternative, we propose translating to and from natural language to perform back-translation between two programming languages. We refer to translating to and from natural language as code summarization and code generation, respectively.
Our proposal is motivated based on the availability of \emph{bimodal} data, source code, and their summaries that are used to train code summarization and generation models.

% In this section, we review the back-translation approach (\cref{sec:bt}), and then describe the multilingual training scheme for code summarization and generation (\cref{sec:sum_gen}). Finally, we detail our approach, summarize--generate to back-translate (\cref{sec:sum_gen_as_bt}).

\subsection{Code Summarization and Generation}
\label{sec:sum_gen}
Source code documentation (\eg docstring or comment) written by software developers is available along with source code on a large scale. Such documentation has been the key source to form code summarization datasets \cite{wan2018improving, hu2018summarizing, leclair-mcmillan-2019-recommendations, husain2019codesearchnet},
% These datasets are also utilized 
and to study natural language (NL) to code generation~\cite{parvez-etal-2021-retrieval-augmented}. 
It is tangible that we can use a code summarization and generation model to translate programming languages. 
% In particular, we envision the natural language as a bridge between programming languages. 
Such a model would first generate an NL summary from an input code in the source language and then generate code from the previously generated NL summary in the target language. As we show in the evaluation, such an approach does not work well in practice (see \cref{table:main_result}); however, code summarization and generation models are viable proxies to generate noisy translations. This enables us to train PLBART, to begin with generating noisy translations, and further learn to improve in a self-supervised fashion when trained via back-translation.
Formally, we jointly train PLBART in a supervised setting to learn code summarization ($\mathcal{S}$) and generation ($\mathcal{G}$):
% using labeled paired data:
\begin{align}
\label{eq:mulsumgen}
\begin{split}
    \mathcal{S} &= TRAIN^{\text{Code} \rightarrow \text{Summary}} \left(\mathcal{P}_{c, s} \right) \\
    \mathcal{G} &= TRAIN^{\text{Summary} \rightarrow \text{Code}} \left(\mathcal{P}_{c, s} \right)
\end{split}
\end{align}
where $\mathcal{P}_{c, s}$ is estimated using the code-to-text benchmark from CodeXGlue \cite{CodeXGLUE}. We follow \citet{tang-etal-2021-multilingual} to perform multilingual fine-tuning of PLBART (in Java and Python) to learn $\mathcal{S}$ and $\mathcal{G}$.
% \footnote{\url{https://github.com/wasiahmad/PLBART/tree/main/multilingual}}
% In the following section, we detail our proposed approach.

\subsection{Back-translation}
\label{sec:bt}
Back-translation (BT) is one of the most popular ways for unsupervised machine translation \cite{artetxe2017unsupervised, lample2018unsupervised, lample-etal-2018-phrase}.
In this approach, we leverage monolingual data in an unsupervised fashion.
% According to the BT approach, 
BT jointly trains a source-to-target model coupled with a backward target-to-source model. 
% trained in parallel
The target-to-source model translates target sequences into the source language, producing noisy sources corresponding to the ground truth target sequences. The source-to-target model is then trained to generate the targets from the noisy sources and vice versa. The two models are trained in parallel until convergence. This training procedure is widely known as \emph{online back-translation} and is the focus of this work.

Back-translation uses a target-to-source model to generate noisy sources and trains a source-to-target model to reconstruct the targets. Specifically, in each step $k$ (a mini-batch update), back-translation performs the following:
\begin{align}
\label{eq:bt}
\begin{split}
    \mathcal{P}_k^{(f)} &= \left\{ (x, f_{k-1} (x)) | x \in \mathcal{D}_{\text{source}} \right\} \\
    b_{k} &= TRAIN^{\text{target} \rightarrow \text{source}} \left(\mathcal{P}_k^{(f)}\right) \\
    \mathcal{P}_k^{(b)} &= \left\{ \left(b_k(y), y\right) | y \in \mathcal{D}_{\text{target}} \right\} \\
    f_{k} &= TRAIN^{\text{source} \rightarrow \text{target}} \left( \mathcal{P}_k^{(b)} \right).
\end{split}
\end{align}
Here, $\mathcal{D}_{source}$, $\mathcal{D}_{target}$ represents unlabeled  data in source and target languages and $TRAIN$ indicates standard sequence-to-sequence training.

Generally, the training via back-translation starts from a forward ($f_0$) and a backward ($b_0$)  model that is trained using parallel data (small gold-standard or large-scale but noisy). Then an extensive collection of unlabeled data is used to train the translation models.
In this work, we assume there is \emph{no} parallel data available across programming languages.
We initialize the forward and backward model with the pre-trained language model, PLBART.
As mentioned before, PLBART cannot generate code in a language different from the input (not even a noisy code) (for example, \cref{fig:qual_ex1}-1).
Therefore, we propose jointly fine-tuning PLBART on code summarization and generation on multiple programming languages in a supervised setting.
Then use the resulting model to initialize the forward and backward model ($f_0$, $b_0$) for back-translation.

\subsection{Summarize--Generate to Back-translate}
\label{sec:sum_gen_as_bt}

% The back-translation method requires the \emph{target-to-source} ($b$) and \emph{source-to-target} ($f$) models to generate semantically equivalent (parallel) sequences in the source and target languages, respectively. 
% These parallel sequences are then used to train the models to learn translations as described in Eq. \eqref{eq:bt}.
% Specifically, the parallel sequences $\{(\hat{x}, y), (x, \hat{y})\}$ created by computing $\hat{y} \leftarrow f_{k-1}(x)$ and $\hat{x} \leftarrow b_k(y)$ kick-start the learning process in back-translation.
% Generally, the curated parallel sequences tend to be noisy since we do not have access to accurate {target-to-source} and {source-to-target} models. 
% However, both the models trained in parallel via back-translation until convergence, resulting in useful translation models.

The recent advancements of pre-trained sequence-to-sequence models on programming languages enables us to use them in initializing the {source-to-target} ($f$) and {target-to-source} ($b$) models for back-translation. Presumably, such pre-trained models should facilitate the learning process during training. 
% Still, they are unable to generate code snippets across languages (as shown in Figure \ref{fig:qual_ex1}) due to their pre-training objective, \ie reconstruction of targets from the noisy sources.
Yet, their pre-training objective -- \ie reconstruction of original input from a noisy source limits their ability to generate code snippets across languages (as shown in \Cref{fig:qual_ex1}).
For example, PLBART as $f(\cdot)$ and $b(\cdot)$ would reconstruct the input, resulting in $f_{k-1}(x) \approx x$ and $b_k(y) \approx y$. 
As a result, the models will learn to merely copy the input sequences rather than translate them.

% In the back-translation method, the \emph{source-to-target} and \emph{target-to-source} models can be initialized from pre-trained sequence-to-sequence models, \eg PLBART. 
% However, due to their inability to generate code across languages (as shown in Figure \ref{fig:qual_ex1}), their direct utilization results in unsuccessful back-translation training. 
% Specifically, PLBART as $f(\cdot)$ and $b(\cdot)$ (shown in Eq. \eqref{eq:bt}) attempts to reconstruct the input, resulting in $f_{k-1}(x) \approx x$ and $b_k(y) \approx y$ that allows the model to copy input rather than to learn translation.
% Therefore, we propose to make use of annotated parallel data between programming and natural languages to fine-tune PLBART and then use its parameters to initialize \emph{source-to-target} and \emph{target-to-source} models for back-translation.

To this end, we propose to make use of available parallel data between programming and natural languages to fine-tune PLBART and then use its parameters to initialize {source-to-target} ($f$) and {target-to-source} ($b$) models for back-translation.
Consequently, we revise the back-translation training method outlined in Eq. \eqref{eq:bt} to follow a \emph{two-step generation} process to perform back-translation: code-to-summary generation in natural language followed by summary-to-code generation in the source language.
Formally, the first $m$ steps (while $k \leq m$) of back-translation is performed as:
\begin{align}
\label{eq:sgb}
\begin{split}
    \mathcal{P}_k^{(f)} &= \left\{ \left(x, {\mathcal{G}\left(\mathcal{S}\left(x\right)\right)}\right) | x \in \mathcal{D}_{\text{source}} \right\} \\
    % b_{k} &= TRAIN^{\text{target} \rightarrow \text{source}} \left(\mathcal{P}_k^{(f)}\right) \\
    \mathcal{P}_k^{(b)} &= \left\{ \left({ \mathcal{G}\left(\mathcal{S}\left(y\right)\right)}, y\right) | y \in \mathcal{D}_{\text{target}} \right\}.
    % \\
    % f_{k} &= TRAIN^{\text{source} \rightarrow \text{target}} \left( \mathcal{P}_k^{(b)} \right).
\end{split}
\end{align}

% The difference between Eq \eqref{eq:bt} and \eqref{eq:sgb} is highlighted in blue.
We find the noisy parallel sequences\footnote{The output sequences are still noisy since the code summarization and generation models are not highly accurate although trained in a supervised fashion.} 
% $\{\left(x, {\color{blue}\mathcal{G}\left(\mathcal{S}\left(x\right)\right)}\right), \left({\color{blue}\mathcal{G}\left(\mathcal{S}\left(x\right)\right)}, y\right)\}$ 
generated via summarization and generation commences the learning process.
% After $m$ steps, back-translation is performed as shown in Eq. \eqref{eq:bt}.
% We find this simple trick allows PLBART to translate target code into ``noisy'' sources that commences the learning process.
The overall idea of our proposed framework is illustrated in Figure \ref{fig:overview} and the Algorithm \ref{alg:sgb} describes the training procedure.
Note that we find it is sufficient to apply our proposed summarization-generation based back-translation only for the first $m$ steps as the {source-to-target} and {target-to-source} models gradually learn to translate, the standard back-translation training reinstated.

\begin{algorithm}[t]
% \renewcommand{\thealgorithm}{}
% \small
\caption{Training Procedure}
{\bf Input:} 
% Parallel data $\mathcal{D}_{(s, nl)}$ and $\mathcal{D}_{(t, nl)}$ between programming languages $s$, $t$ and natural language $nl$; 
Monolingual (unlabeled) data $\mathcal{D}_{source}$ and $\mathcal{D}_{target}$; number of initial steps $m$; number of total steps $I$; code summarizer $\mathcal{S}(\cdot,\cdot)$; code generator $\mathcal{G}(\cdot,\cdot)$; parameters $\theta$ to initialize the forward and backward translation models $f(\cdot,\cdot)$ and $b(\cdot,\cdot)$. \\
{\bf Output:} Final model parameters $\theta$.
\begin{algorithmic}[1]
% \Require
\For{$k = 0,\cdots,I$}
    % \State $y_s \leftarrow$ Sample a batch from $\mathcal{D}_s$
    % \State $y_t \leftarrow$ Sample a batch from $\mathcal{D}_t$
    \State $y \leftarrow (y_s \sim\mathcal{D}_{source}) \cup (y_t \sim\mathcal{D}_{target}$)
    \If{$k \leq m$}
        % \State $\hat{x}_s \sim \mathcal{G}_{\theta}(\cdot|\mathcal{S}_{\theta}( \cdot|y_t))$\
        % \State $\hat{x}_t \sim \mathcal{G}_{\theta}(\cdot|\mathcal{S}_{\theta}(\cdot|y_s))$
        \State $x_{nl} \sim \mathcal{S}(\cdot|y)$ \Comment{code-to-summary}
        \State $\hat{x} \sim \mathcal{G}(\cdot|x_{nl})$
        \Comment{summary-to-code}
    \Else
        % \State $\hat{x}_s \sim b_{\theta}(\cdot | y_t)$
        % \State $\hat{x}_t \sim f_{\theta}(\cdot | y_s)$
        \State $\hat{x} \leftarrow (x_s \sim b(\cdot|y_t)) \cup (x_t \sim f(\cdot|y_s))$ 
        % \Comment{Back-translation}
    \EndIf
    % \State $\mathcal{B} \leftarrow \{\langle \hat{x}_s, y_t\rangle\} \cup \{\langle \hat{x}_t, y_s\rangle\}$
    \State Update $\theta$ by maximizing log-likelihood of $f(\hat{x}_s, y_t)$ and $b(\hat{x}_t, y_s)$
    % and $b_{\theta}(\hat{x}_t, y_s)$
\EndFor
\end{algorithmic}
\label{alg:sgb}
\end{algorithm}

% In this work, we apply our proposed framework to build translation system in Java and Python programming languages.

% \paragraph{Python-to-Java translation}
% \begin{align}
% \begin{split}
%      \mathcal{P}_k^{(\mathcal{S},\ \mathcal{G})} &= \left\{ \left(\mathcal{G}\left(\mathcal{S}\left(y\right)\right), y\right)\ |\ y \in \mathcal{D}_{\text{Java}} \right\} \\
%     b_{k} &= TRAIN^{\text{Python} \rightarrow \text{Java}} \left( \mathcal{P}_k^{(\mathcal{S},\ \mathcal{G})} \right).
% \end{split}
% \end{align}

% \paragraph{Java-to-Python translation}
% \begin{align}
% \begin{split}
%     \mathcal{P}_k^{(\mathcal{S},\ \mathcal{G})} &= \left\{ \left(\mathcal{G}\left(\mathcal{S}\left(y\right)\right), y\right)\ |\ y \in \mathcal{D}_{\text{Python}} \right\} \\
%     f_{k} &= TRAIN^{\text{Java} \rightarrow \text{Python}} \left( \mathcal{P}_k^{(\mathcal{S},\ \mathcal{G})} \right).
% \end{split}
% \end{align}

\section{Experiment Setup}

\subsection{Models and Baselines}

\paragraph{Our model} 
% As noted earlier, PLBART \cite{ahmad-etal-2021-unified} and CodeT5 \cite{wang-etal-2021-codet5} are two popular sequence-to-sequence models pre-trained on source code that cannot be trained via back-translation (BT). As an alternative, our proposed approach can be leveraged to train both models to learn programming language translation in an unsupervised fashion. In this work, we chose PLBART to perform experiments and show the effectiveness of our proposed approach.
Our proposed approach can be applied to pre-trained sequence-to-sequence models, \eg PLBART \cite{ahmad-etal-2021-unified} and CodeT5 \cite{wang-etal-2021-codet5}. In this work, we chose PLBART\footnote{Since its pretraining implementation is publicly available at \url{https://github.com/wasiahmad/PLBART}.} to perform experiments and show the effectiveness of our proposed framework.

\subsection*{Baseline Models}
We compare our proposed approach applied to PLBART with the following existing approaches.
 
% \smallskip
% \noindent$\bullet$~\textbf{j2py \hspace{0.5em}}
\paragraph{j2py} 
is a framework that translates Java source code to Python.\footnote{\url{https://github.com/natural/java2python}} 
It follows handwritten rules manually built using expert knowledge.

% \smallskip
% \noindent$\bullet$~\textbf{Summarize-and-Generate ($\mathcal{S} \& \mathcal{G}$) \hspace{0.5em}}
\paragraph{Summarize-and-Generate ($\mathcal{S} \& \mathcal{G}$)}
performs code-to-code translation via two steps, code-to-summary and summary-to-code generation. We evaluate the $\mathcal{S} \& \mathcal{G}$ model (as in Eq. \eqref{eq:mulsumgen}) that is used to perform code summarization and generation in our proposed framework to train PLBART via BT.

% \noindent$\bullet$~\textbf{TransCoder}
\paragraph{TransCoder} 
% \smallskip
% \noindent$\bullet$~\textbf{TransCoder \hspace{0.5em}} 
is a neural translation model for programming languages \cite{lachaux2020unsupervised}. TransCoder is developed by pretraining Transformer \cite{vaswani2017attention} via masked language modeling (MLM) objective \cite{devlin2018bert} on monolingual source code datasets. In a second step, TransCoder is trained via denoising autoencoding (DAE) and BT. In this work, we consider TransCoder as the \textbf{primary} baseline.\footnote{We compare TransCoder and PLBART in terms of model architecture and training setup in the Appendix D.}

% \notewa{We should mention the differences between PLBART and TransCoder to highlight, why their comparison is not fair.}

\paragraph{DOBF}
% \smallskip
% \noindent$\bullet$~\textbf{DOBF \hspace{0.5em}} 
\citet{roziere2021dobf} proposed a pretraining objective, DOBF, that leverages the structural aspects of programming languages. According to this pretraining paradigm, the identifiers (class, function, and variable names) in code snippets are obfuscated, and a model is trained to recover the original names. DOBF shares the same neural architecture as TransCoder. We report the evaluation performances of TransCoder and DOBF from the official code release by \citet{lachaux2020unsupervised}.\footnote{ \url{https://github.com/facebookresearch/CodeGen/blob/main/docs/transcoder.md\#results}).}

\begin{table}[t]
\centering
% \resizebox{\linewidth}{!}
{%
% \small
\begin{tabular}{l r r}
\hline
 & Java & Python \\ 
\hline
\multicolumn{3}{l}{Github - \emph{unimodal} data} \\
\cdashline{1-3}
% Nb of functions & 44.5 M & 42.0 M \\
% Nb of tokens & 3.3 B & 4.1 B \\
Nb of functions & 7.2 M & 8.3 M \\
Nb of tokens & 752 M & 665 M \\
\hline
\multicolumn{3}{l}{CodeNet - \emph{unimodal} data} \\
\cdashline{1-3}
Nb of functions & 0.42 M & 0.15 M \\
Nb of tokens & 47.3 M & 17.0 M \\
\hline
\multicolumn{3}{l}{CodeXGlue - \emph{bimodal} data} \\
\cdashline{1-3}
Nb of functions & 164,923 & 251,818 \\
Nb of tokens & 21.2 M & 44.3 M \\
\hline
\end{tabular}
}
% \vspace{-2mm}
\caption{
Statistics of the data used to train PLBART at different stages in this work. \emph{Bimodal} data refers to parallel function-summary pairs, while \emph{unimodal} data refers to monolingual (and unparallel) functions.
}
\label{table:data_stat}
% \vspace{-2mm}
\end{table}

\subsection{Evaluation Dataset and Metrics}

\paragraph{Evaluation Dataset}
\citet{lachaux2020unsupervised} proposed an evaluation dataset composed of parallel functions in Java, Python, and C++ languages. The dataset consists of 464 Java to Python and 482 Python to Java test examples, with 10 unit test cases accompanying each.

\subsection*{Evaluation Metrics}

\paragraph{BLEU}
% \noindent$\bullet$~\textbf{BLEU \hspace{0.5em}} 
measures n-gram overlap between a generated translation and a collection of reference translations \cite{papineni-etal-2002-bleu}.

% \smallskip
% \noindent$\bullet$~\textbf{Exact Match (EM) \hspace{0.5em}} 
\paragraph{Exact Match (EM)} 
represents the percentage of generated translations exactly match with the collection of reference translations. 

% \smallskip
% \noindent$\bullet$~\textbf{CodeBLEU \hspace{0.5em}} 
\paragraph{CodeBLEU}
% \notewa{adding definition of CodeBLEU will probably help.}
measures grammatical and logical correctness in addition to n-gram overlap between generated and reference translations \cite{ren2020codebleu}. 
% CodeBLEU assesses grammatical and logical correctness based on the abstract syntax tree and the data-flow structure.
CodeBLEU is defined as a weighted sum of n-gram match, weighted n-gram match,\footnote{ Different weights are assigned to n-grams such that the keywords (\eg \texttt{for}, \texttt{while}) have higher weights} syntax match (based on AST), and data-flow match.

% \smallskip
% \noindent$\bullet$~\textbf{Computational Accuracy (CA) \hspace{0.5em}} 
\paragraph{Computational Accuracy (CA),} proposed by \citet{lachaux2020unsupervised}, assess functional correctness; a translated code is considered correct if it passes a set of unit tests. 
It evaluates if a generated function outputs the same as the reference when given the same set of inputs. This metric overcomes the shortcoming of match-based metrics (\eg BLEU, CodeBLEU) by accounting for the program-execution behavior \cite{lachaux2020unsupervised,chen2021evaluating}. 
% Several recent works \cite{lachaux2020unsupervised,chen2021evaluating} showed that match-based metrics (\eg BLEU, CodeBLEU) are unable to account for the large and complex space of programs functionally equivalent to a reference solution.
% \citet{lachaux2020unsupervised} introduced this evaluation
%  to assess functional correctness; a translated code is considered correct if it passes a set of unit tests.
% perform evaluation based on unit tests. 

\begin{table*}[!ht]
\centering
\resizebox{\linewidth}{!}
{%
% \small%
% \def\arraystretch{1.05}%
\begin{tabular}{l|c c c c|c c c c}
% \begin{tabular}{@{}l@{\hskip 0.05in} | c@{\hskip 0.05in} c@{\hskip 0.05in} c@{\hskip 0.1in} c@{\hskip 0.1in} | c@{\hskip 0.05in} c@{\hskip 0.05in} c@{\hskip 0.1in} c@{\hskip 0.1in}}
\hline
\multirow{2}{*}{{Models}} & \multicolumn{4}{c|}{{Java $\rightarrow$ Python}} & \multicolumn{4}{c}{{Python $\rightarrow$ Java}} \\ 
\cline{2-9}
& BLEU  & EM & CodeBLEU & CA & BLEU & EM & CodeBLEU & CA  \\ 
\hline
j2py* & - & - & - & 38.3 & - & - & - & - \\
% TransCoder$^\ast$ & 68.1 & 3.7 & - & 35.0 & 64.6 & 0.8 & - & 24.7 \\
TransCoder$^\ast$ & 68.1 & 3.7 & - & 46.9 & 64.6 & 0.8 & - & 32.6 \\
TransCoder w/ DOBF$^\ast$ & - & - & - & 49.2 & - & - & - & 40.4 \\
% \cdashline{1-9}
\hdashline
\sng \eqref{eq:mulsumgen} & 7.6 & 0.0 & 15.8 & 0.2 & 12.4 & 0 & 16.3 & 0.2 \\
% TransCoder & 67.6 & 1.9 & 65.6 & 47.0 & 65.0 & 0.6 & 70.3 & 32.6 \\
% DOBF & {\bf 71.4} & {\bf 3.5} & {\bf 67.4} & {\bf 49.1} & {\bf 66.3} & {\bf 1.8} & {\bf 70.4} & 39.6 \\
\cdashline{1-9}
\multicolumn{9}{l}{PLBART \bf{(this work)}} \\
\cdashline{1-9}
\hdashline
trained via BT & 31.2 & 0.0 & 36.6 & 0.0 & 31.7 & 0.0 & 32.1 & 0.0 \\
trained via BT (via \sng) & 64.2 & 2.8 & 63.4 & 40.4 & 64.1 & 2.1 & 65.9 & 31.9 \\
\hline
\end{tabular}
}
% \vspace{-2mm}
\caption{
Evaluation results of the baselines models and our proposed framework using greedy decoding. $^\ast$ indicates the updated scores reported in the official code repository of \citet{lachaux2020unsupervised}. 
Note that, TransCoder and PLBART models have 312M and 140M parameters, respectively.
}
\label{table:main_result}
% \vspace{-2mm}
\end{table*}

\subsection{Training Datasets and Preprocessing}
\label{subsec:dataset}

\paragraph{Code Summarization and Generation} \citet{CodeXGLUE} curated a code summarization dataset consisting of code and summary pairs based on the CodeSearchNet dataset \cite{husain2019codesearchnet}. We use this dataset in Java and Python programming languages to train the code-to-summary and summary-to-code generation models.

\paragraph{Back-translation (BT)}

For BT training (as discussed in \cref{sec:sum_gen_as_bt}), we use the GitHub public dataset available on Google BigQuery \cite{hoffa2016github}.\footnote{\url{https://console.cloud.google.com/marketplace/product/github/github-repos}} 
We first deduplicate\footnote{We used a hash-based data deduplication method.} the GitHub dataset at the program level, extract the functions, and finally perform another deduplication at the function level. 
% This dataset is also used to pre-train PLBART and TransCoder.
Note that the Github dataset is composed of source code that covers a wide variety of programming topics (as they come from various projects). In contrast, the evaluation dataset is composed of programming problems covering basic data structure and algorithmic concepts. Therefore, to investigate the impact of data on BT training, we alternatively chose \emph{unparallel} code samples in Java and Python from CodeNet \cite{puri2021project}. The CodeNet dataset is collected from two online judge websites, \emph{AIZU Online Judge} and \emph{AtCoder}, and composed of submissions for 4053 problems. We use the deduplicated accepted solutions to the problems for BT training. Presumably, CodeNet and the evaluation dataset \cite{lachaux2020unsupervised} have a similar nature that should positively impact downstream translation performance.

% \footnote{\url{https://onlinejudge.u-aizu.ac.jp}} and AtCoder\footnote{\url{https://atcoder.jp}}.

% \paragraph{Github}
% \noindent$\bullet$~\textbf{Github} 

% \smallskip
% \noindent$\bullet$~\textbf{CodeNet} 
% \paragraph{CodeNet} 

\paragraph{Preprocessing}
We use \texttt{tree\_sitter}\footnote{\url{https://github.com/tree-sitter}} for tokenizing Java functions and the tokenizer of the standard library for Python.\footnote{\url{https://docs.python.org/3/library/tokenize.html}}
We extract standalone functions\footnote{Standalone functions can be used without instantiating a class. In Java, this corresponds to static methods, and in Python, it corresponds to functions outside classes.} from the BT training datasets following the function extraction technique from \citet{lachaux2020unsupervised}.
Considering our computational budget, we filter the standalone functions exceeding a maximum length of 256 to cope with our computational resources.
The statistics of the preprocessed datasets are presented in Table \ref{table:data_stat}.

% For preprocessing source code, we use the publicly available source code of TransCoder \cite{lachaux2020unsupervised}.\footnote{https://github.com/facebookresearch/CodeGen}

\subsection{Implementation Details}
We jointly train PLBART on code summarization and generation in Java and Python using the authors' provided code.\footnote{\url{https://github.com/wasiahmad/PLBART/tree/main/multilingual}}
Subsequently, we train PLBART via back-translation as described in Algorithm \ref{alg:sgb}. We set $I = 10,000$ and tuned $m = 200$.\footnote{We tuned $m$ in the range [100, 1000] with 100 steps.}
% proposed in this work for a maximum of 10,000 steps with 100 warm-up steps.
We train PLBART using 8 Nvidia GeForce RTX 2080 Ti GPUs, and the effective batch size is maintained at 1024 instances at both training stages. 
We optimize PLBART with the Adam optimizer \cite{kingma2014adam}, a learning rate of 10e-4, and use a polynomial learning rate decay scheduling.
The best models are selected based on the validation BLEU scores.
We implement our approach in Fairseq~\cite{ott-etal-2019-fairseq} and use float16 operations to speed up training.

% \footnote{\url{https://github.com/wasiahmad/PLBART}}

\paragraph{Decoding}
During inference, we use beam search decoding \cite{koen-2004-pharaoh} to generate multiple translations using PLBART. We chose greedy search (Beam 1) as the \emph{default} decoding scheme for validation and evaluation.
However, following \citet{lachaux2020unsupervised}, we report two sets of results for the computational accuracy (CA) metric: CA@n B=n, the percentage of functions with at least one correct translation in the beam (of size n), and CA@1 B=n the percentage of functions where the hypothesis in the beam with the highest log-probability is a correct translation.

\section{Results and Analysis}
% We aim to address the following questions.
% \begin{compactenum}
%     \item Does our proposed approach empowers PLBART to learn programming language translation? (\cref{sec:main_result})
%     \item Does summarization and generation enable us to create parallel examples to warmup PLBART for back-translation? (\cref{sec:analysis1})
%     \item Does the use of in-domain data (CodeNet) in training lead to better translations? (\cref{sec:analysis2})
% \end{compactenum}

% \begin{table}[!ht]
% \centering
% \resizebox{\linewidth}{!}
% {%
% % \small%
% % \def\arraystretch{1.05}%
% \begin{tabular}{l|c@{\hskip 0.1in} c}
% \hline
% Models & Java $\rightarrow$ Python & Python $\rightarrow$ Java \\ 
% \hline
% \multicolumn{3}{l}{\emph{TransCoder}} \\
% \cdashline{1-3}
% CA@1 k=1 & 35.0 & 35.5 \\
% CA@1 k=10 & 49.0 & 38.4 \\
% CA@5 k=5 & 60.0 & 47.0 \\
% CA@10 k=10 & 64.4 & 50.0 \\
% \cdashline{1-3}
% \multicolumn{3}{l}{\emph{PLBART}} \\
% \cdashline{1-3}
% CA@1 k=1 & 24.7 & 43.8 \\
% CA@1 k=10 & 36.6 & 45.4 \\
% CA@5 k=5 & 44.3 & 52.5 \\
% CA@10 k=10 & 51.1 & 55.0 \\
% \hline
% \end{tabular}
% }
% % \vspace{-2mm}
% \caption{
% Write caption.
% }
% \label{table:beam_result}
% % \vspace{-2mm}
% \end{table}

\begin{table}[!ht]
\centering
% \resizebox{\linewidth}{!}
{%
% \small%
% \def\arraystretch{1.05}%
\begin{tabular}{l|c@{\hskip 0.1in} c}
\hline
Models & TransCoder & PLBART \\ 
\hline
\multicolumn{3}{l}{\emph{Java $\rightarrow$ Python}} \\
\cdashline{1-3}
CA@1 B=1 & 46.9 & 40.4 \\
CA@1 B=10 & 48.8 & 41.8 \\
CA@5 B=5 & 60.0 & 47.7 \\
CA@10 B=10 & 64.4 & 50.3 \\
\cdashline{1-3}
\multicolumn{3}{l}{\emph{Python $\rightarrow$ Java}} \\
\cdashline{1-3}
CA@1 B=1 & 32.6 & 31.9 \\
CA@1 B=10 & 36.0 & 34.5 \\
CA@5 B=5 & 44.3 & 45.1 \\
CA@10 B=10 & 51.1 & 50.0 \\
\hline
\end{tabular}
}
% \vspace{-2mm}
\caption{
Computational accuracy (CA@m) with beam search decoding and comparison between TransCoder and PLBART. TransCoder's performances are reported from \citet{lachaux2020unsupervised}. 
The value B indicates the beam size. 
CA@m B=n means that we use beam decoding to generate n translations, and select the top m translations based on their log-probability scores. 
}
\label{table:beam_result}
% \vspace{-2mm}
\end{table}

\subsection{Main Result}
\label{sec:main_result}

\Cref{table:main_result} shows the performance of our proposed approach and the baseline models on both Java to Python and Python to Java translation. 
We begin by comparing PLBART directly used in back-translation (BT) training with our proposed approach (the last block in \Cref{table:main_result}). 
Since PLBART does not know to generate across languages, when the model is trained via BT, it only learns to copy the input sources. As a result, PLBART scores 0\% EM and 0\% CA, while 30+ BLEU and CodeBLEU scores.
In contrast, following our proposed approach of summarizing and generating to back-translate, PLBART trained via BT (via \sng) achieves 40.4\% and 31.9\% CA scores.
This performance is competitive to state-of-the-art translation system, TransCoder.\footnote{Note that, while comparing PLBART with TransCoder on the translation performance, their differences (shown in \Cref{table:replication_params}) should be taken into consideration.}
We compare them using beam search decoding in \Cref{table:beam_result}.

\begin{table*}[!ht]
\centering
% \resizebox{\linewidth}{!}
{%
% \small%
% \def\arraystretch{1.05}%
\begin{tabular}{l|c c c c|c c c c}
\hline
\multirow{ 2}{*}{Approach} & \multicolumn{4}{c|}{{Java to Python}} & \multicolumn{4}{c}{{Python to Java}} \\ 
\cline{2-9}
& BLEU  & EM & CodeBLEU & CA & BLEU & EM & CodeBLEU & CA  \\ 
\hline
Warm-start w/ PD & 60.5 & 2.8 & 61.1 & { 41.9} & 62.6 & { 2.4} & 65.9 & 32.0 \\
Proposed approach & { 64.2} & 2.8 & { 63.4} & 40.4 & { 64.1} & 2.1 & 65.9 & 31.9 \\
\hline
\end{tabular}
}
\vspace{-2mm}
\caption{
Comparison between PLBART warm-started using parallel data (PD) and our approach to summarize and generate to back-translate on the fly during the initial steps of back-translation training.
% Evaluation results when pre-trained PLBART checkpoint and ``PLBART-MM'' checkpoint is used as the starting point for back-translation based training. ``PLBART-MM'' refers to the model checkpoint resulted from multilingual code summarization and generation training (it is the default setting in this work).
}
\label{table:checkpoint_impact}
% \vspace{-2mm}
\end{table*}

\begin{table*}[!ht]
\centering
% \resizebox{\linewidth}{!}
{%
% \small%
% \def\arraystretch{1.05}%
\begin{tabular}{l|c c c c|c c c c}
\hline
\multirow{ 2}{*}{{Data Source}} & \multicolumn{4}{c|}{{Java to Python}} & \multicolumn{4}{c}{{Python to Java}} \\ 
\cline{2-9}
& BLEU  & EM & CodeBLEU & CA & BLEU & EM & CodeBLEU & CA  \\ 
\hline
Github & 64.2 & 2.8 & 63.4 & 40.4 & 64.1 & 2.1 & 65.9 & 31.9 \\
CodeNet & 65.6 & 3.1 & 64.7 & 50.9 & 65.1 & 2.5 & 68.5 & 46.5 \\
\hline
\end{tabular}
}
% \vspace{-2mm}
\caption{
PLBART evaluation results when our proposed framework uses data from Github (available via BigQuery \cite{hoffa2016github}) and competitive programming sites (available via CodeNet \cite{puri2021project}).
}
\label{table:data_impact}
% \vspace{-2mm}
\end{table*}

% The primary objective of this work is to study the feasibility of using pre-trained sequence-to-sequence models for unsupervised programming language translation via BT. 
% The experimental results conclude that such models cannot directly be used in BT training; however, training via \sng empowers them to generate across languages and further be trained via BT to learn to translate.
Overall, the experimental results confirm our conjecture that pre-trained sequence-to-sequence models cannot be effectively used in BT training; however, training via \sng empowers them to generate across languages and be further trained via BT to learn programming language translation.

% indicate that they are not a reliable indicator of translation correctness.
% \footnote{Several recent works \cite{kulal2019spoc,lachaux2020unsupervised,chen2021evaluating} showed that match-based metrics (\eg BLEU, CodeBLEU) are unable to account for the large and complex space of programs functionally equivalent to a reference solution.} 
% \notewa{Can we clarify further?}

% Although PLBART is a multi-lingual model pre-trained to generate code, it does not work well when directly trained with back translation, as evident by the result (0\% EM and 0\% CA). We attribute such poor performance to PLBART's lack of knowledge about cross-lingual alignment. pre-trained Multi-Lingual Sequence to Sequence Source Code Models such as PLBART, as is, cannot translate across languages without further training.
% We also show the performance of a PLBART model trained via Back-Translation.
% Finally, we show the empirical results of the PLBART model, which is first trained to Summarize and Generate and then trained via Back-Translation (BT). 

% Note that, PLBART lags comparing to DOBF on Java to Python translation as DOBF has shown to perform better on downstream tasks due to its pre-training objective.
% The results suggest that while \sng and PLBART trained via BT does not produce useful translations, PLBART trained via BT initialized from \sng capable to perform translation.

\subsection{Analysis}

\paragraph{Summarize and generate to create parallel data}
% \label{sec:analysis1}
Our proposed approach generates parallel code sequences on the fly (online) for training. An alternative to our approach is to use a code summarization and generation model to create parallel code sequences (offline) and warm-start PLBART for back-translation-based training. We compare these two approaches in \Cref{table:checkpoint_impact}, and the results show that both approaches perform comparably.
However, it is essential to note that the online setting gives us flexibility as we can tune the number of initial steps ($m$ in Algorithm \ref{alg:sgb}).
In contrast, the offline setting requires generating a sufficiently large number of parallel code sequences for effective training.

% \repeatit[5]{Placeholder\par}

\paragraph{Impact of in-domain training data}
% \label{sec:analysis2}
The evaluation dataset comprises solutions to programming problems involving data structures and algorithm concepts. While Github offers large-scale unlabeled data, most of its code belongs to software projects that use APIs and advanced functionalities. Therefore, we utilize an alternative dataset called CodeNet collected from two online judge websites. We refer to this dataset as \emph{in-domain} since its nature aligns with the evaluation dataset (data structure and algorithm focused problems aggregated from GeeksforGeeks).
We compare in-domain data usage with Github data on BT-based training.
The results in \Cref{table:data_impact} show that the use of in-domain data significantly boosts the performance in both translation directions.
A detailed error analysis reveals that such a performance boost is due to a reduction in {\tt TypeError}. 
We speculate that in-domain data have similarities in the data type usage that helps the model.
% Due to the page limit, we present more findings of the error analysis and qualitative examples in the Appendix.
We present further error analysis and qualitative examples in the Appendix.
% In addition, we discuss the limitations and risks of using our proposed model in the Appendix.
% In addition, we perform a rigorous error analysis to show the strength and weaknesses of PLBART-based translation model. 

% \input{figures/example1}

% \paragraph{Qualitative Examples}
% \Cref{fig:example1} shows an example of Java-to-Python translation by PLBART. The translated code is both syntactically and semantically correct \ie our compiler could successfully parse and build the translated code. It passed 2 test cases out of 10 when executed. The translated code is slightly different from the input Java code. In particular, line 13 in the input Java code is a loop that iterates backward (decreasing order). However, line 12 in the generated python code iterates forward (increasing order). 
% If the generated python code was {\tt range(c-1, 0, -1)} instead of {\tt range(c-1)}, it would pass all the test cases. 
% We attribute such behavior to the fact that {\tt range(*)} is much more frequent pattern than {\tt range(*, 0, -1)} in python code.

% We provide more analysis in Appendix.

\section{Related Work}

% \subsection{Programming Language Translation}
\paragraph{Programming Language Translation}
% \noindent\textbf{Programming Language Translation \hspace{0.5em}}
Translating programs or source code across different programming languages (PL) requires a profound understanding of the PLs. Having strictly defined syntax and semantics, PLs are suitable for phrase-based statistical machine translation~\cite{nguyen2013lexical, karaivanov2014phrase, aggarwal2015using}. 
\citet{chen2018tree} introduced a tree-to-tree machine translation to translate programs and to learn the syntactic alignment between source and target PL. Recently proposed pre-trained programming language models showed promising results in translating programs across PLs~\cite{feng2020codebert, guo2020graphcodebert, ahmad-etal-2021-unified, ahmad2021avatar}. However, these approaches require a set of parallel programs to train the encoder-decoder model.

Recently proposed Transcoder \cite{lachaux2020unsupervised} shows initial success results in unsupervised program translation, eliminating the requirement of bi-modal data. They achieve such jointly training a model using XLM~\cite{NIPS2019_8928}, Denoising Auto Encoding (DAE)~\cite{vincent2008extracting}, and Back-Translation(BT)~\cite{lample2018unsupervised}. This work empirically investigated the suitability of adopting BT to train existing pre-trained encoder-decoder models and proposed an alternative via summarization and generation.

% \notewa{We want to discuss prior works as in Section 2 of this paper \cite{roziere2021leveraging} on ``Translation of Programming Languages''.}

% \subsection{Unsupervised Machine Translation via Back-translation}

% \vspace{-4pt}
\paragraph{Unsupervised Machine Translation via Back-translation}
% \smallskip
% \noindent\textbf{Unsupervised Machine Translation via Back-translation \hspace{0.5em}}
Gathering sufficiently large parallel corpora has been a significant challenge for Machine Translation (MT)~\cite{guzman-etal-2019-flores}. Several research efforts are invested in learning MT using monolingual data~\cite{artetxe-etal-2018-unsupervised, artetxe2017unsupervised, lachaux2020unsupervised} to solve this problem. For example, \citet{gulcehre2015using} proposed integrating a Language model into the decoder. \citet{xia2016dual} proposed Neural MT (NMT) as a bidirectional and dual learning task. More recent advancements in unsupervised MT leverages back-translation (BT)~\cite{sennrich-etal-2016-improving, lample2018unsupervised, lample-etal-2018-phrase}. 
In back-translation, the target-to-source model generates noisy sources given target sequences and then trains the source-to-target model to reconstruct the targets and vice versa. 
While BT has been widely adopted for unsupervised NMT, it is used in other applications \cite{zhu2017unpaired, pmlr-v80-hoffman18a, shen2017style, yang2018unsupervised, zhang2018style}.

% in cycle-consistency \cite{zhu2017unpaired, pmlr-v80-hoffman18a} and style transfer \cite{shen2017style, yang2018unsupervised, zhang2018style}.

\section{Conclusion}
In this research, we show that pre-trained sequence-to-sequence models ({\em e.g.}, PLBART) are not suitable for direct adaptation via back-translation to learn to translate. 
% Due to the lack of knowledge about cross-lingual generation, such models cannot generate code in the source language given an input code in the target language.
% , which is a crucial step in back-translation. 
% As a solution to this problem, we propose to use code summarization and generation as an alternative to target-to-source translation in the back-translation approach. 
To address the issue, we propose to use code summarization and generation as an alternative to performing back-translation. 
% We leverage \emph{bimodal} data, e.g., source code and their summaries, to train a code-to-summary and a summary-to-code generation model. Subsequently, these models are used to generate target-to-summary and summary-to-source, which is equivalent to target-to-source translation, enabling training pre-trained programming language models via back-translation.
We show that our proposed approach turns PLBART into a translation model that performs competitively to existing unsupervised translation models.

\section*{Limitations}
One of the risks of using our developed translation model is that we used the Github dataset for training that may contain information that uniquely identifies an individual or offensive content.
Since we are developing the translation model for research purposes \emph{only}, we believe our usage of the Github data does not violate their licensing terms and conditions.
While we do not present it as a justification, the PLBART model was pre-trained on the Github data that may include sensitive information.
As we converted PLBART into a programming language translation model, it is unlikely to generate sensitive information unless it is provided such information as input. However, we should be careful while using translation models trained using unfiltered data.
All the experiments performed in this work are based only on one seed. Therefore, using other random seeds may lead to results that could be different from ours.

\section*{Ethics Statement}
\paragraph{Training data and its risks} 
We use the GitHub public dataset available on Google BigQuery filtered to keep only projects with open-source licenses.\footnote{We select the open-source licenses: `apache-2.0', `mit', `gpl-2.0', `gpl-3.0', `bsd-2-clause', `bsd-3-clause'.}
While we do not perform preprocessing that would eliminate any personally identifiable information or offensive content, we remove natural language comments that presumably reduce toxic content. 
Nonetheless, using code language models (LMs) comes with certain risks, e.g., generating biased, toxic, and vulnerable code. \citet{chen2021evaluating} discussed the broader impact and risks of code LMs (Section 7). We should keep those factors to ensure the responsible use of code LMs.

\paragraph{Carbon Footprint} We avoided using large models, reducing their environmental impacts. We train \texttt{PLBART-base} model on summarization-generation and backtranslation for a maximum of 10k steps on 8 \texttt{RTX 2080 Ti} GPUs that took 1-2 days. Therefore, the training would emit approximately 15kg of carbon into the environment.\footnote{Calculated using \url{https://mlco2.github.io/impact}, based on a total of 200 hours of training on RTX 2080 Ti and Amazon Web Services as the provider.} No model finetuning is performed in this work.

\bibliography{anthology,custom}

\begin{thebibliography}{49}
\expandafter\ifx\csname natexlab\endcsname\relax\def\natexlab#1{#1}\fi

\bibitem[{Aggarwal et~al.(2015)Aggarwal, Salameh, and
  Hindle}]{aggarwal2015using}
Karan Aggarwal, Mohammad Salameh, and Abram Hindle. 2015.
\newblock \href {https://peerj.com/preprints/1459.pdf} {Using machine
  translation for converting python 2 to python 3 code}.
\newblock Technical report, PeerJ PrePrints.

\bibitem[{Ahmad et~al.(2021{\natexlab{a}})Ahmad, Chakraborty, Ray, and
  Chang}]{ahmad-etal-2021-unified}
Wasi Ahmad, Saikat Chakraborty, Baishakhi Ray, and Kai-Wei Chang.
  2021{\natexlab{a}}.
\newblock \href {https://doi.org/10.18653/v1/2021.naacl-main.211} {Unified
  pre-training for program understanding and generation}.
\newblock In \emph{Proceedings of the 2021 Conference of the North American
  Chapter of the Association for Computational Linguistics: Human Language
  Technologies}, pages 2655--2668, Online. Association for Computational
  Linguistics.

\bibitem[{Ahmad et~al.(2021{\natexlab{b}})Ahmad, Tushar, Chakraborty, and
  Chang}]{ahmad2021avatar}
Wasi~Uddin Ahmad, Md~Golam~Rahman Tushar, Saikat Chakraborty, and Kai-Wei
  Chang. 2021{\natexlab{b}}.
\newblock \href {https://arxiv.org/abs/2108.11590} {Avatar: A parallel corpus
  for java-python program translation}.
\newblock \emph{arXiv preprint arXiv:2108.11590}.

\bibitem[{Artetxe et~al.(2018{\natexlab{a}})Artetxe, Labaka, and
  Agirre}]{artetxe-etal-2018-unsupervised}
Mikel Artetxe, Gorka Labaka, and Eneko Agirre. 2018{\natexlab{a}}.
\newblock \href {https://doi.org/10.18653/v1/D18-1399} {Unsupervised
  statistical machine translation}.
\newblock In \emph{Proceedings of the 2018 Conference on Empirical Methods in
  Natural Language Processing}, pages 3632--3642, Brussels, Belgium.
  Association for Computational Linguistics.

\bibitem[{Artetxe et~al.(2018{\natexlab{b}})Artetxe, Labaka, Agirre, and
  Cho}]{artetxe2017unsupervised}
Mikel Artetxe, Gorka Labaka, Eneko Agirre, and Kyunghyun Cho.
  2018{\natexlab{b}}.
\newblock \href {https://openreview.net/pdf?id=Sy2ogebAW} {Unsupervised neural
  machine translation}.
\newblock In \emph{International Conference on Learning Representations}.

\bibitem[{Artetxe and Schwenk(2019)}]{artetxe-schwenk-2019-massively}
Mikel Artetxe and Holger Schwenk. 2019.
\newblock \href {https://doi.org/10.1162/tacl_a_00288} {Massively multilingual
  sentence embeddings for zero-shot cross-lingual transfer and beyond}.
\newblock \emph{Transactions of the Association for Computational Linguistics},
  7:597--610.

\bibitem[{Bahdanau et~al.(2015)Bahdanau, Cho, and Bengio}]{bahdanau2014neural}
Dzmitry Bahdanau, Kyunghyun Cho, and Yoshua Bengio. 2015.
\newblock \href {https://arxiv.org/abs/1409.0473} {Neural machine translation
  by jointly learning to align and translate}.
\newblock In \emph{International Conference on Learning Representations}.

\bibitem[{Chen et~al.(2021)Chen, Tworek, Jun, Yuan, Pinto, Kaplan, Edwards,
  Burda, Joseph, Brockman et~al.}]{chen2021evaluating}
Mark Chen, Jerry Tworek, Heewoo Jun, Qiming Yuan, Henrique Ponde de~Oliveira
  Pinto, Jared Kaplan, Harri Edwards, Yuri Burda, Nicholas Joseph, Greg
  Brockman, et~al. 2021.
\newblock Evaluating large language models trained on code.
\newblock \emph{arXiv preprint arXiv:2107.03374}.

\bibitem[{Chen et~al.(2018)Chen, Liu, and Song}]{chen2018tree}
Xinyun Chen, Chang Liu, and Dawn Song. 2018.
\newblock \href
  {http://papers.nips.cc/paper/7521-tree-to-tree-neural-networks-for-program-translation.pdf}
  {Tree-to-tree neural networks for program translation}.
\newblock In \emph{Advances in Neural Information Processing Systems 31}, pages
  2547--2557. Curran Associates, Inc.

\bibitem[{Conneau and Lample(2019)}]{NIPS2019_8928}
Alexis Conneau and Guillaume Lample. 2019.
\newblock \href
  {http://papers.nips.cc/paper/8928-cross-lingual-language-model-pretraining.pdf}
  {Cross-lingual language model pretraining}.
\newblock In H.~Wallach, H.~Larochelle, A.~Beygelzimer, F.~d\textquotesingle
  Alch\'{e}-Buc, E.~Fox, and R.~Garnett, editors, \emph{Advances in Neural
  Information Processing Systems 32}, pages 7059--7069. Curran Associates, Inc.

\bibitem[{Devlin et~al.(2019)Devlin, Chang, Lee, and
  Toutanova}]{devlin2018bert}
Jacob Devlin, Ming-Wei Chang, Kenton Lee, and Kristina Toutanova. 2019.
\newblock \href {https://doi.org/10.18653/v1/N19-1423} {{BERT}: Pre-training of
  deep bidirectional transformers for language understanding}.
\newblock In \emph{Proceedings of the 2019 Conference of the North {A}merican
  Chapter of the Association for Computational Linguistics: Human Language
  Technologies, Volume 1 (Long and Short Papers)}, pages 4171--4186,
  Minneapolis, Minnesota. Association for Computational Linguistics.

\bibitem[{Ding et~al.(2022)Ding, Buratti, Pujar, Morari, Ray, and
  Chakraborty}]{ding2021contrastive}
Yangruibo Ding, Luca Buratti, Saurabh Pujar, Alessandro Morari, Baishakhi Ray,
  and Saikat Chakraborty. 2022.
\newblock \href {https://doi.org/10.18653/v1/2022.acl-long.436} {Towards
  learning (dis)-similarity of source code from program contrasts}.
\newblock In \emph{Proceedings of the 60th Annual Meeting of the Association
  for Computational Linguistics (Volume 1: Long Papers)}, pages 6300--6312,
  Dublin, Ireland. Association for Computational Linguistics.

\bibitem[{Edunov et~al.(2018)Edunov, Ott, Auli, and
  Grangier}]{edunov-etal-2018-understanding}
Sergey Edunov, Myle Ott, Michael Auli, and David Grangier. 2018.
\newblock \href {https://doi.org/10.18653/v1/D18-1045} {Understanding
  back-translation at scale}.
\newblock In \emph{Proceedings of the 2018 Conference on Empirical Methods in
  Natural Language Processing}, pages 489--500, Brussels, Belgium. Association
  for Computational Linguistics.

\bibitem[{Feng et~al.(2020{\natexlab{a}})Feng, Guo, Tang, Duan, Feng, Gong,
  Shou, Qin, Liu, Jiang, and Zhou}]{feng-etal-2020-codebert}
Zhangyin Feng, Daya Guo, Duyu Tang, Nan Duan, Xiaocheng Feng, Ming Gong, Linjun
  Shou, Bing Qin, Ting Liu, Daxin Jiang, and Ming Zhou. 2020{\natexlab{a}}.
\newblock \href {https://doi.org/10.18653/v1/2020.findings-emnlp.139}
  {{C}ode{BERT}: A pre-trained model for programming and natural languages}.
\newblock In \emph{Findings of the Association for Computational Linguistics:
  EMNLP 2020}, pages 1536--1547, Online. Association for Computational
  Linguistics.

\bibitem[{Feng et~al.(2020{\natexlab{b}})Feng, Guo, Tang, Duan, Feng, Gong,
  Shou, Qin, Liu, Jiang, and Zhou}]{feng2020codebert}
Zhangyin Feng, Daya Guo, Duyu Tang, Nan Duan, Xiaocheng Feng, Ming Gong, Linjun
  Shou, Bing Qin, Ting Liu, Daxin Jiang, and Ming Zhou. 2020{\natexlab{b}}.
\newblock \href {https://www.aclweb.org/anthology/2020.findings-emnlp.139}
  {{C}ode{BERT}: A pre-trained model for programming and natural languages}.
\newblock In \emph{Findings of the Association for Computational Linguistics:
  EMNLP 2020}, pages 1536--1547, Online. Association for Computational
  Linguistics.

\bibitem[{Gulcehre et~al.(2015)Gulcehre, Firat, Xu, Cho, Barrault, Lin,
  Bougares, Schwenk, and Bengio}]{gulcehre2015using}
Caglar Gulcehre, Orhan Firat, Kelvin Xu, Kyunghyun Cho, Loic Barrault, Huei-Chi
  Lin, Fethi Bougares, Holger Schwenk, and Yoshua Bengio. 2015.
\newblock \href {https://arxiv.org/abs/1503.03535} {On using monolingual
  corpora in neural machine translation}.
\newblock \emph{arXiv preprint arXiv:1503.03535}.

\bibitem[{Guo et~al.(2021)Guo, Ren, Lu, Feng, Tang, Liu, Zhou, Duan, Yin, Jiang
  et~al.}]{guo2020graphcodebert}
Daya Guo, Shuo Ren, Shuai Lu, Zhangyin Feng, Duyu Tang, Shujie Liu, Long Zhou,
  Nan Duan, Jian Yin, Daxin Jiang, et~al. 2021.
\newblock \href {https://openreview.net/forum?id=jLoC4ez43PZ} {Graphcodebert:
  Pre-training code representations with data flow}.
\newblock In \emph{International Conference on Learning Representations}.

\bibitem[{Guzm{\'a}n et~al.(2019)Guzm{\'a}n, Chen, Ott, Pino, Lample, Koehn,
  Chaudhary, and Ranzato}]{guzman-etal-2019-flores}
Francisco Guzm{\'a}n, Peng-Jen Chen, Myle Ott, Juan Pino, Guillaume Lample,
  Philipp Koehn, Vishrav Chaudhary, and Marc{'}Aurelio Ranzato. 2019.
\newblock \href {https://doi.org/10.18653/v1/D19-1632} {The {FLORES} evaluation
  datasets for low-resource machine translation: {N}epali{--}{E}nglish and
  {S}inhala{--}{E}nglish}.
\newblock In \emph{Proceedings of the 2019 Conference on Empirical Methods in
  Natural Language Processing and the 9th International Joint Conference on
  Natural Language Processing (EMNLP-IJCNLP)}, pages 6098--6111, Hong Kong,
  China. Association for Computational Linguistics.

\bibitem[{He et~al.(2016)He, Xia, Qin, Wang, Yu, Liu, and Ma}]{xia2016dual}
Di~He, Yingce Xia, Tao Qin, Liwei Wang, Nenghai Yu, Tie-Yan Liu, and Wei-Ying
  Ma. 2016.
\newblock \href
  {https://proceedings.neurips.cc/paper/2016/file/5b69b9cb83065d403869739ae7f0995e-Paper.pdf}
  {Dual learning for machine translation}.
\newblock In \emph{Advances in Neural Information Processing Systems},
  volume~29. Curran Associates, Inc.

\bibitem[{Hoffa(2016)}]{hoffa2016github}
Felipe Hoffa. 2016.
\newblock \href
  {https://cloud.google.com/blog/topics/public-datasets/github-on-bigquery-analyze-all-the-open-source-code}
  {Github on bigquery: Analyze all the open source code}.

\bibitem[{Hoffman et~al.(2018)Hoffman, Tzeng, Park, Zhu, Isola, Saenko, Efros,
  and Darrell}]{pmlr-v80-hoffman18a}
Judy Hoffman, Eric Tzeng, Taesung Park, Jun-Yan Zhu, Phillip Isola, Kate
  Saenko, Alexei Efros, and Trevor Darrell. 2018.
\newblock \href {https://proceedings.mlr.press/v80/hoffman18a.html}
  {{C}y{CADA}: Cycle-consistent adversarial domain adaptation}.
\newblock In \emph{Proceedings of the 35th International Conference on Machine
  Learning}, volume~80 of \emph{Proceedings of Machine Learning Research},
  pages 1989--1998. PMLR.

\bibitem[{Hu et~al.(2018)Hu, Li, Xia, Lo, Lu, and Jin}]{hu2018summarizing}
Xing Hu, Ge~Li, Xin Xia, David Lo, Shuai Lu, and Zhi Jin. 2018.
\newblock \href {https://doi.org/10.24963/ijcai.2018/314} {Summarizing source
  code with transferred api knowledge}.
\newblock In \emph{Proceedings of the Twenty-Seventh International Joint
  Conference on Artificial Intelligence, {IJCAI-18}}, pages 2269--2275.
  International Joint Conferences on Artificial Intelligence Organization.

\bibitem[{Husain et~al.(2019)Husain, Wu, Gazit, Allamanis, and
  Brockschmidt}]{husain2019codesearchnet}
Hamel Husain, Ho-Hsiang Wu, Tiferet Gazit, Miltiadis Allamanis, and Marc
  Brockschmidt. 2019.
\newblock \href {https://arxiv.org/abs/1909.09436} {Codesearchnet challenge:
  Evaluating the state of semantic code search}.
\newblock \emph{arXiv preprint arXiv:1909.09436}.

\bibitem[{Karaivanov et~al.(2014)Karaivanov, Raychev, and
  Vechev}]{karaivanov2014phrase}
Svetoslav Karaivanov, Veselin Raychev, and Martin Vechev. 2014.
\newblock \href {https://doi.org/10.1145/2661136.2661148} {Phrase-based
  statistical translation of programming languages}.
\newblock In \emph{Proceedings of the 2014 ACM International Symposium on New
  Ideas, New Paradigms, and Reflections on Programming \& Software}, pages
  173--184.

\bibitem[{Kingma and Ba(2015)}]{kingma2014adam}
Diederik~P. Kingma and Jimmy Ba. 2015.
\newblock \href {http://arxiv.org/abs/1412.6980} {Adam: {A} method for
  stochastic optimization}.
\newblock In \emph{3rd International Conference on Learning Representations,
  {ICLR} 2015, San Diego, CA, USA, May 7-9, 2015, Conference Track
  Proceedings}.

\bibitem[{Koen(2004)}]{koen-2004-pharaoh}
Philipp Koen. 2004.
\newblock \href
  {https://link.springer.com/chapter/10.1007/978-3-540-30194-3_13} {Pharaoh: a
  beam search decoder for phrase-based statistical machine translation models}.
\newblock In \emph{Proceedings of the 6th Conference of the Association for
  Machine Translation in the Americas: Technical Papers}, pages 115--124,
  Washington, USA. Springer.

\bibitem[{Lachaux et~al.(2020)Lachaux, Roziere, Chanussot, and
  Lample}]{lachaux2020unsupervised}
Marie-Anne Lachaux, Baptiste Roziere, Lowik Chanussot, and Guillaume Lample.
  2020.
\newblock \href
  {https://proceedings.neurips.cc/paper/2020/file/ed23fbf18c2cd35f8c7f8de44f85c08d-Paper.pdf}
  {Unsupervised translation of programming languages}.
\newblock In \emph{Advances in Neural Information Processing Systems},
  volume~33, pages 20601--20611. Curran Associates, Inc.

\bibitem[{Lample et~al.(2018{\natexlab{a}})Lample, Conneau, Denoyer, and
  Ranzato}]{lample2018unsupervised}
Guillaume Lample, Alexis Conneau, Ludovic Denoyer, and Marc'Aurelio Ranzato.
  2018{\natexlab{a}}.
\newblock \href {https://openreview.net/forum?id=rkYTTf-AZ} {Unsupervised
  machine translation using monolingual corpora only}.
\newblock In \emph{International Conference on Learning Representations}.

\bibitem[{Lample et~al.(2018{\natexlab{b}})Lample, Ott, Conneau, Denoyer, and
  Ranzato}]{lample-etal-2018-phrase}
Guillaume Lample, Myle Ott, Alexis Conneau, Ludovic Denoyer, and Marc{'}Aurelio
  Ranzato. 2018{\natexlab{b}}.
\newblock \href {https://doi.org/10.18653/v1/D18-1549} {Phrase-based {\&}
  neural unsupervised machine translation}.
\newblock In \emph{Proceedings of the 2018 Conference on Empirical Methods in
  Natural Language Processing}, pages 5039--5049, Brussels, Belgium.
  Association for Computational Linguistics.

\bibitem[{LeClair and McMillan(2019)}]{leclair-mcmillan-2019-recommendations}
Alexander LeClair and Collin McMillan. 2019.
\newblock \href {https://doi.org/10.18653/v1/N19-1394} {Recommendations for
  datasets for source code summarization}.
\newblock In \emph{Proceedings of the 2019 Conference of the North {A}merican
  Chapter of the Association for Computational Linguistics: Human Language
  Technologies, Volume 1 (Long and Short Papers)}, pages 3931--3937,
  Minneapolis, Minnesota. Association for Computational Linguistics.

\bibitem[{Lu et~al.(2021)Lu, Guo, Ren, Huang, Svyatkovskiy, Blanco, Clement,
  Drain, Jiang, Tang et~al.}]{CodeXGLUE}
Shuai Lu, Daya Guo, Shuo Ren, Junjie Huang, Alexey Svyatkovskiy, Ambrosio
  Blanco, Colin Clement, Dawn Drain, Daxin Jiang, Duyu Tang, et~al. 2021.
\newblock \href {https://arxiv.org/abs/2102.04664} {Codexglue: A machine
  learning benchmark dataset for code understanding and generation}.
\newblock \emph{arXiv preprint arXiv:2102.04664}.

\bibitem[{Nguyen et~al.(2013)Nguyen, Nguyen, and Nguyen}]{nguyen2013lexical}
Anh~Tuan Nguyen, Tung~Thanh Nguyen, and Tien~N Nguyen. 2013.
\newblock \href {https://doi.org/10.1145/2491411.2494584} {Lexical statistical
  machine translation for language migration}.
\newblock In \emph{Proceedings of the 2013 9th Joint Meeting on Foundations of
  Software Engineering}, pages 651--654.

\bibitem[{Niu et~al.(2022)Niu, Li, Ng, Ge, Huang, and Luo}]{niu2022spt}
Changan Niu, Chuanyi Li, Vincent Ng, Jidong Ge, Liguo Huang, and Bin Luo. 2022.
\newblock Spt-code: Sequence-to-sequence pre-training for learning the
  representation of source code.
\newblock In \emph{2022 IEEE/ACM 44th International Conference on Software
  Engineering (ICSE)}. Association for Computing Machinery.

\bibitem[{Ott et~al.(2019)Ott, Edunov, Baevski, Fan, Gross, Ng, Grangier, and
  Auli}]{ott-etal-2019-fairseq}
Myle Ott, Sergey Edunov, Alexei Baevski, Angela Fan, Sam Gross, Nathan Ng,
  David Grangier, and Michael Auli. 2019.
\newblock \href {https://doi.org/10.18653/v1/N19-4009} {fairseq: A fast,
  extensible toolkit for sequence modeling}.
\newblock In \emph{Proceedings of the 2019 Conference of the North {A}merican
  Chapter of the Association for Computational Linguistics (Demonstrations)},
  pages 48--53, Minneapolis, Minnesota. Association for Computational
  Linguistics.

\bibitem[{Papineni et~al.(2002)Papineni, Roukos, Ward, and
  Zhu}]{papineni-etal-2002-bleu}
Kishore Papineni, Salim Roukos, Todd Ward, and Wei-Jing Zhu. 2002.
\newblock \href {https://doi.org/10.3115/1073083.1073135} {{B}leu: a method for
  automatic evaluation of machine translation}.
\newblock In \emph{Proceedings of the 40th Annual Meeting of the Association
  for Computational Linguistics}, pages 311--318, Philadelphia, Pennsylvania,
  USA. Association for Computational Linguistics.

\bibitem[{Parvez et~al.(2021)Parvez, Ahmad, Chakraborty, Ray, and
  Chang}]{parvez-etal-2021-retrieval-augmented}
Md~Rizwan Parvez, Wasi Ahmad, Saikat Chakraborty, Baishakhi Ray, and Kai-Wei
  Chang. 2021.
\newblock \href {https://doi.org/10.18653/v1/2021.findings-emnlp.232}
  {Retrieval augmented code generation and summarization}.
\newblock In \emph{Findings of the Association for Computational Linguistics:
  EMNLP 2021}, pages 2719--2734, Punta Cana, Dominican Republic. Association
  for Computational Linguistics.

\bibitem[{Puri et~al.(2021)Puri, Kung, Janssen, Zhang, Domeniconi, Zolotov,
  Dolby, Chen, Choudhury, Decker et~al.}]{puri2021project}
Ruchir Puri, David~S Kung, Geert Janssen, Wei Zhang, Giacomo Domeniconi,
  Vladmir Zolotov, Julian Dolby, Jie Chen, Mihir Choudhury, Lindsey Decker,
  et~al. 2021.
\newblock \href
  {https://datasets-benchmarks-proceedings.neurips.cc/paper/2021/file/a5bfc9e07964f8dddeb95fc584cd965d-Paper-round2.pdf}
  {Project codenet: A large-scale ai for code dataset for learning a diversity
  of coding tasks}.
\newblock In \emph{Proceedings of the Neural Information Processing Systems
  Track on Datasets and Benchmarks}, volume~1.

\bibitem[{Ren et~al.(2020)Ren, Guo, Lu, Zhou, Liu, Tang, Zhou, Blanco, and
  Ma}]{ren2020codebleu}
Shuo Ren, Daya Guo, Shuai Lu, Long Zhou, Shujie Liu, Duyu Tang, Ming Zhou,
  Ambrosio Blanco, and Shuai Ma. 2020.
\newblock \href {https://arxiv.org/abs/2009.10297} {Codebleu: a method for
  automatic evaluation of code synthesis}.
\newblock \emph{arXiv preprint arXiv:2009.10297}.

\bibitem[{Roziere et~al.(2021)Roziere, Lachaux, Szafraniec, and
  Lample}]{roziere2021dobf}
Baptiste Roziere, Marie-Anne Lachaux, Marc Szafraniec, and Guillaume Lample.
  2021.
\newblock \href {https://arxiv.org/abs/2102.07492} {Dobf: A deobfuscation
  pre-training objective for programming languages}.
\newblock In \emph{Advances in Neural Information Processing Systems}.

\bibitem[{Sennrich et~al.(2016)Sennrich, Haddow, and
  Birch}]{sennrich-etal-2016-improving}
Rico Sennrich, Barry Haddow, and Alexandra Birch. 2016.
\newblock \href {https://doi.org/10.18653/v1/P16-1009} {Improving neural
  machine translation models with monolingual data}.
\newblock In \emph{Proceedings of the 54th Annual Meeting of the Association
  for Computational Linguistics (Volume 1: Long Papers)}, pages 86--96, Berlin,
  Germany. Association for Computational Linguistics.

\bibitem[{Shen et~al.(2017)Shen, Lei, Barzilay, and Jaakkola}]{shen2017style}
Tianxiao Shen, Tao Lei, Regina Barzilay, and Tommi Jaakkola. 2017.
\newblock \href
  {https://papers.nips.cc/paper/2017/file/2d2c8394e31101a261abf1784302bf75-Paper.pdf}
  {Style transfer from non-parallel text by cross-alignment}.
\newblock In \emph{Advances in Neural Information Processing Systems 30}.

\bibitem[{Tang et~al.(2021)Tang, Tran, Li, Chen, Goyal, Chaudhary, Gu, and
  Fan}]{tang-etal-2021-multilingual}
Yuqing Tang, Chau Tran, Xian Li, Peng-Jen Chen, Naman Goyal, Vishrav Chaudhary,
  Jiatao Gu, and Angela Fan. 2021.
\newblock \href {https://doi.org/10.18653/v1/2021.findings-acl.304}
  {Multilingual translation from denoising pre-training}.
\newblock In \emph{Findings of the Association for Computational Linguistics:
  ACL-IJCNLP 2021}, pages 3450--3466, Online. Association for Computational
  Linguistics.

\bibitem[{Vaswani et~al.(2017)Vaswani, Shazeer, Parmar, Uszkoreit, Jones,
  Gomez, Kaiser, and Polosukhin}]{vaswani2017attention}
Ashish Vaswani, Noam Shazeer, Niki Parmar, Jakob Uszkoreit, Llion Jones,
  Aidan~N Gomez, \L~ukasz Kaiser, and Illia Polosukhin. 2017.
\newblock \href
  {http://papers.nips.cc/paper/7181-attention-is-all-you-need.pdf} {Attention
  is all you need}.
\newblock In \emph{Advances in Neural Information Processing Systems 30}, pages
  5998--6008. Curran Associates, Inc.

\bibitem[{Vincent et~al.(2008)Vincent, Larochelle, Bengio, and
  Manzagol}]{vincent2008extracting}
Pascal Vincent, Hugo Larochelle, Yoshua Bengio, and Pierre-Antoine Manzagol.
  2008.
\newblock \href {https://doi.org/10.1145/1390156.1390294} {Extracting and
  composing robust features with denoising autoencoders}.
\newblock In \emph{Proceedings of the 25th international conference on Machine
  learning}, pages 1096--1103.

\bibitem[{Wan et~al.(2018)Wan, Zhao, Yang, Xu, Ying, Wu, and
  Yu}]{wan2018improving}
Yao Wan, Zhou Zhao, Min Yang, Guandong Xu, Haochao Ying, Jian Wu, and Philip~S.
  Yu. 2018.
\newblock \href {https://doi.org/10.1145/3238147.3238206} {Improving automatic
  source code summarization via deep reinforcement learning}.
\newblock In \emph{Proceedings of the 33rd ACM/IEEE International Conference on
  Automated Software Engineering}, ASE 2018, page 397–407, New York, NY, USA.
  Association for Computing Machinery.

\bibitem[{Wang et~al.(2021)Wang, Wang, Joty, and Hoi}]{wang-etal-2021-codet5}
Yue Wang, Weishi Wang, Shafiq Joty, and Steven~C.H. Hoi. 2021.
\newblock \href {https://doi.org/10.18653/v1/2021.emnlp-main.685} {{C}ode{T}5:
  Identifier-aware unified pre-trained encoder-decoder models for code
  understanding and generation}.
\newblock In \emph{Proceedings of the 2021 Conference on Empirical Methods in
  Natural Language Processing}, pages 8696--8708, Online and Punta Cana,
  Dominican Republic. Association for Computational Linguistics.

\bibitem[{Yang et~al.(2018)Yang, Hu, Dyer, Xing, and
  Berg-Kirkpatrick}]{yang2018unsupervised}
Zichao Yang, Zhiting Hu, Chris Dyer, Eric~P Xing, and Taylor Berg-Kirkpatrick.
  2018.
\newblock \href {https://dl.acm.org/doi/pdf/10.5555/3327757.3327831}
  {Unsupervised text style transfer using language models as discriminators}.
\newblock In \emph{Proceedings of the 32nd International Conference on Neural
  Information Processing Systems}, pages 7298--7309.

\bibitem[{Zhang et~al.(2019)Zhang, Ren, Liu, Wang, Chen, Li, Zhou, and
  Chen}]{zhang2018style}
Zhirui Zhang, Shuo Ren, Shujie Liu, Jianyong Wang, Peng Chen, Mu~Li, Ming Zhou,
  and Enhong Chen. 2019.
\newblock \href {https://arxiv.org/abs/1808.07894} {Style transfer as
  unsupervised machine translation}.
\newblock In \emph{Thirty-Third AAAI Conference on Artificial Intelligence}.

\bibitem[{Zhu et~al.(2017)Zhu, Park, Isola, and Efros}]{zhu2017unpaired}
Jun-Yan Zhu, Taesung Park, Phillip Isola, and Alexei~A Efros. 2017.
\newblock \href
  {https://openaccess.thecvf.com/content_ICCV_2017/papers/Zhu_Unpaired_Image-To-Image_Translation_ICCV_2017_paper.pdf}
  {Unpaired image-to-image translation using cycle-consistent adversarial
  networks}.
\newblock In \emph{Proceedings of the IEEE international conference on computer
  vision}, pages 2223--2232.

\end{thebibliography}
\bibliographystyle{acl_natbib}

\clearpage
\appendix
\twocolumn[{%
 \centering
 \Large\bf Supplementary Material: Appendices \\ [20pt]
}]

\begin{table}[!ht]
\centering
% \resizebox{\linewidth}{!}
{%
% \small%
% \def\arraystretch{1.05}%
\begin{tabular}{l |c c}
\hline
& TransCoder & PLBART \\
\hline
\multicolumn{3}{l}{\emph{Java $\rightarrow$ Python}} \\
\cdashline{1-3}
\#Tests & 464 & 464 \\
Error & 149 & 146 \\
Failure & 93 & 123 \\
Success & 218 & 188 \\
\quad EM & 17  & 24 \\
Timeout & 4 & 7 \\
\cdashline{1-3}
\multicolumn{3}{l}{\emph{Python $\rightarrow$ Java}} \\
\cdashline{1-3}
\#Tests & 482 & 482 \\
Error & 201 & 212 \\
Failure & 118 & 108 \\
Success & 157 & 154 \\
\quad EM & 6 & 2 \\
Timeout & 6 & 8 \\
\hline
\end{tabular}
}
% \vspace{-2mm}
\caption{
Detailed results of computational accuracy using greedy decoding for Java $\leftrightarrow$ Python translation. 
}
\label{table:comp_acc}
% \vspace{-2mm}
\end{table}

\section{Analysis of Computational Accuracy}
\label{sec:comp_anal}

\Cref{table:comp_acc} shows the breakdown of computational accuracies for Java-to-Python and Python-to-Java translation for TransCoder and our proposed approach using PLBART. 
We execute the generated function and match the output \wrt the expected output. 
TransCoder results in 149 error cases, 93 failure cases, and 218 success cases in Java-to-Python translation, with 17 solutions matching the ground truth. In contrast, PLBART results in 146 error cases, 123 failure cases, and 188 success cases. Out of these 188 successes in PLBART, 24 solutions exactly match the target solution. 

For Python-Java translation, TransCoder results in 201 errors, 118 failures, and 157 successes, out of which 6 are an exact match. On the other hand, in the case of PLBART, there are 212 error cases, 108 failure cases, and 154 success cases, out of which two exactly match the target solution. Performing human study to understand why translated functions fail test cases would facilitate model comparisons, and we leave it as future work.

\begin{table}[!ht]
\centering
\resizebox{\linewidth}{!}
{%
% \small%
% \def\arraystretch{1.05}%
\begin{tabular}{l@{\hskip 0.1in} c@{\hskip 0.1in} c}
\hline
Error Category & TransCoder & PLBART  \\ 
\midrule
\rowcolor{dark-gray}
\#Errors (Java $\rightarrow$ Python)  & 149 & 146 \\
\midrule
\rowcolor{light-gray}
Compilation & - & - \\
\midrule
\rowcolor{light-gray}
Runtime & 149 & 146 \\
\quad TypeError  & 47 & 61 \\
\quad IndexError  & 18 & 20 \\
\quad NameError  & 17 & 16 \\
\quad ValueError  & 11 & 15 \\
\quad UnboundLocalError & 13 & 11 \\
\quad Others & 17 & 14 \\
\quad SyntaxError & 26 & 9 \\
\midrule
\rowcolor{dark-gray}
\#Errors (Python $\rightarrow$ Java) & 201 & 212 \\
\midrule
\rowcolor{light-gray}
Compilation & 151 & 180 \\
\quad TypeError  & 89 & 108 \\
\quad CantFindSymbol  & 23 & 30 \\
\quad SyntaxError  & 14 & 25 \\
\quad BadOperand  & 15 & 12 \\
\quad Others & 10 & 5 \\
\midrule
\rowcolor{light-gray}
Runtime & 50 & 27 \\
\quad IndexOutOfBoundsE.  & 40 & 15 \\
\quad NumberFormatE.  & 5 & 6 \\
\quad NullPointerE.  & 2 & 3 \\
\quad Others & 3 & 3 \\
\hline
\end{tabular}
}
% \vspace{-2mm}
\caption{
Category of errors made by the TransCoder and PLBART translation models. 
The error categories are sorted based on the PLBART's error count on the respective category. 
In Python $\rightarrow$ Java runtime error categories, ``E.'' stands for ``Exception''.
% A detailed categorization is provided in the Appendix.
}
\label{table:errors_made_by_model}
% \vspace{-2mm}
\end{table}

\begin{table*}[!ht]
\centering
% \resizebox{\linewidth}{!}
% {%
% \small%
% \def\arraystretch{1.05}%
\begin{tabular}{p{11cm} c}
\hline
Error Category & Count  \\ 
\midrule
\rowcolor{light-gray}
Type Error & 61 \\
list indices must be integers or slices, not {\bf A} & 18 \\
{\bf A} object does not support item assignment & 13 \\
{\bf A} object cannot be interpreted as an integer & 8 \\
unsupported/bad operand type(s) & 10 \\
int object is not iterable/callable/subscriptable & 6 \\
Others & 6 \\
\midrule
\rowcolor{light-gray}
Index Error & 20 \\
{\bf B} index out of range & 19 \\
others & 1 \\
\midrule
\rowcolor{light-gray}
Name Error & 16 \\
name {\bf C} is not defined & 16 \\
\midrule
\rowcolor{light-gray}
Value Error & 15 \\
not enough values to unpack & 7 \\
too many values to unpack & 3 \\
the truth value of an array with more than one element is ambiguous & 3 \\
others & 2 \\
\midrule
\rowcolor{light-gray}
Unbound Local Error & 11 \\
local variable {\bf D} referenced before assignment & 11 \\
\hline
\end{tabular}
% }
% \vspace{-2mm}
\caption{
Analyzing the five most frequent error cases (123 out of 146) encountered in PLBART generated Java to Python translation.
{\bf A} and {\bf B} indicate \{bool, int, tuple, str, range\} and \{string, list\}, respectively. {\bf C} and {\bf D} indicate identifier (class, function, variable) names.
}
\label{table:error_analysis_plbart_top_errors}
% \vspace{-2mm}
\end{table*}

\begin{figure*}[!htb]
% \rule{\linewidth}{0.6pt}
\centering
\begin{adjustbox}{valign=t,minipage=0.47\textwidth}
% \underline{Input function in Java} \\
\begin{center}
    \underline{\large Input function in Java}
\end{center}
\begin{tabular}{l}
\lstset{escapechar=~,style=CustomJava}
\begin{lstlisting}[ 
    linebackgroundcolor={%
    \ifnum\value{lstnumber}=13
        \color{green!10}
    \fi
    }
]
static int sortedCount(int mat[][], 
            int r, int c){  
    int result = 0;  
    for(int i = 0; i < r; i++){                 
        int j;   
        for(j = 0; j < c - 1; j++) 
            if(mat[i][j + 1] <= mat[i][j]) 
                break; 
        if(j == c - 1) result++;
    }        
    for(int i = 0; i < r; i++){
        int j; 
        for(j = c - 1; j > 0; j--) 
            if(mat[i][j - 1] <= mat[i][j]) 
                break;
        if(c > 1 && j == 0) result++;
    }
    return result;
}
\end{lstlisting}
\end{tabular}
\end{adjustbox}
\hspace{10pt}
% \vspace{10pt}
\begin{adjustbox}{valign=t,minipage=0.47\textwidth}
% \underline{Translated function in Python} \\
\begin{center}
    \underline{\large Translated function in Python}
\end{center} 
\begin{tabular}{l}
\lstset{escapechar=@,style=CustomPy}
\begin{lstlisting}[ 
    linebackgroundcolor={%
    \ifnum\value{lstnumber}=12
            \color{red!10}
    \fi
    }
]
def sorted_count(mat, r, c):      
    result = 0
    for i in range(r):
        j = 0
        for j in range(c - 1):
            if mat[i][j + 1] <= mat[i][j]:
                break
        if j == c - 1:
            result += 1
    for i in range(r):
        j = c - 1
        for j in range(c - 1):
            if mat[i][j - 1] <= mat[i][j]:
                break
        if c > 1 and j == 0:
            result += 1
    return result
\end{lstlisting}
\end{tabular}
\end{adjustbox}
% \vspace{-2mm}
\caption{
An example of Java to Python translation by PLBART that passes {\bf 2 out of 10} unit test cases. Line no. 13 (marked in {green}) in the Java function is incorrectly translated in python (line no. 12, marked in {red}). Replacing the \emph{range} function parameter ``(c-1)'' by ``(c - 1, 0, -1)'' would make the translated function pass all the test cases.
% problem link: https://www.geeksforgeeks.org/count-sorted-rows-matrix
}
% \vspace{-2mm}
\label{fig:example1}
\end{figure*}

% \begin{table}[!tbh]
% \centering
% % \resizebox{\linewidth}{!}
% {%
% % \small%
% % \def\arraystretch{1.05}%
% \begin{tabular}{l c c}
% \hline
%  & TransCoder & PLBART  \\
% \hline
% \#layers* & 6 & 6 \\
% \#heads & 8 & 12 \\
% Model dim & 1024 & 768 \\
% Vocab size & 64,000 & 50,000 \\
% Total parameters  & 312 M & 140 M \\
% \cline{1-3}
% \multicolumn{3}{l}{\textbf{{Stage1: Pre-training}}} \\ 
% \cline{1-3}
% Objective & MLM & DAE \\
% Total tokens & 920 B & 87 B \\
% Token types & BPE & Sentencepiece \\
% \hdashline
% \multirow{3}{*}{Languages} & Java, & Java,\\
% & Python, & Python, \\
% & C++ & English\\
% \hline
% \multicolumn{3}{l}{\textbf{{Stage2: Training}}} \\ 
% \cline{1-3}
% Objective & DAE+BT & BT \\
% Total tokens & 625 M & 430 M \\
% Token types & BPE & Sentencepiece \\
% \hdashline
% \multirow{3}{*}{Languages} & Java, & Java,\\
% & Python, & Python, \\
% & C++ & \\

% \hline
% \end{tabular}
% \\ * \#layers represent both encoder and decoder number of layers.
% }
% \caption{
% Training and parameters setting of Transcoder and out approach.
% }
% \label{table:replication_params}
% % \vspace{-2mm}
% \end{table}

\begin{table*}[!ht]
\centering
% \resizebox{\linewidth}{!}
{%
% \small%
% \def\arraystretch{1.05}%
\begin{tabular}{l c c}
\hline
 & TransCoder & PLBART  \\
\hline
\#layers (encoder) & 6 & 6 \\
\#layers (decoder) & 6 & 6 \\
\#heads & 8 & 12 \\
Model dim & 1024 & 768 \\
Vocab size & 64,000 & 50,000 \\
Total parameters  & 312 M & 140 M \\
\cdashline{1-3}
\multicolumn{3}{l}{{Stage1: Pre-training}} \\ 
\cdashline{1-3}
Objective & MLM & DAE \\
Total tokens & 920 B & 87 B \\
Token types & BPE & Sentencepiece \\
Languages & Java, Python, C++ & Java, Python, English \\
\cdashline{1-3}
\multicolumn{3}{l}{{Stage2: Training}} \\ 
\cdashline{1-3}
Objective & DAE+BT & BT \\
Total tokens & 625 M & 430 M \\
Token types & BPE & Sentencepiece \\
Languages & Java, Python, C++ & Java, Python \\
\hline
\end{tabular}
}
\caption{
TransCoder vs. PLBART.
}
\label{table:replication_params}
% \vspace{-2mm}
\end{table*}

% \begin{table}[!ht]
% \centering
% \resizebox{\linewidth}{!}
% {%
% % \small%
% % \def\arraystretch{1.05}%
% \begin{tabular}{l c@{\hskip 0.1in} c}
% \hline
%  & TransCoder & PLBART  \\
% \hline
% \#Enc-layers & 6 & 6 \\
% \#Dec-layers & 6 & 6 \\
% \#Heads & 8 & 12 \\
% Model dim & 1024 & 768 \\
% Vocab size & 64,000 & 50,000 \\
% Total parameters  & 312 M & 140 M \\
% \cdashline{1-3}
% \multicolumn{3}{l}{{Stage1: Pre-training}} \\ 
% \cdashline{1-3}
% Objective & MLM & DAE \\
% Total tokens & 920 B & 87 B \\
% Token types & BPE & Sentencepiece \\
% Languages & \makecell{Java, Python, \\ C++} & \makecell{Java, Python, \\ English} \\
% \cdashline{1-3}
% \multicolumn{3}{l}{{Stage2: Training}} \\ 
% \cdashline{1-3}
% Objective & DAE+BT & BT \\
% Total tokens & 625 M & 430 M \\
% Token types & BPE & Sentencepiece \\
% Languages & \makecell{Java, Python,\\ C++} & Java, Python \\
% \hline
% \end{tabular}
% }
% \caption{
% TransCoder vs. PLBART.
% }
% \label{table:replication_params}
% % \vspace{-2mm}
% \end{table}

% To understand the quality of errors, we 
% Compilation vs. Runtime Errors \ref{fig:comp_vs_runtime}.

\section{Error Analysis}
\label{sec:error_anal}
We further analyze the error cases for TransCoder and our proposed approach using PLBART. Since Python is an interpreted language, syntactic and semantic errors are caught at runtime. Thus, we categorize all errors for Java-to-Python translation as runtime errors. \Cref{table:errors_made_by_model} shows the errors in both Java-to-Python and Python-to-Java translation. 
While PLBART is susceptible to {\tt TypeError}, TransCoder is disproportionately susceptible to {\tt SyntaxError}. 
In the case of Python-to-Java translation, PLBART exhibits more Compilation errors, but TransCoder exhibits more Runtime errors. 
The most common compilation error type in TransCoder and PLBART is {\tt TypeError}. The most common runtime error in Python-to-Java translation is {\tt InderOutOfBoundException} for both models, where TransCoder exhibits more than twice the number of such errors in PLBART.

Finally, we identified the top five error categories (which account for 123 errors out of 146) exhibited by PLBART in Java-to-Python translation and analyzed the error messages. In most cases, {\tt TypeError} and {\tt ValueError} are due to a mismatch in the underlying data type of variable. \Cref{table:error_analysis_plbart_top_errors} shows the detailed statistics of different error types, sub-types, and their frequencies. As mentioned earlier, training using in-domain data collected from CodeNet \cite{puri2021project} significantly reduces {\tt TypeError}. We hypothesize that in-domain data have similarities in the data type usage that helps the model improve. 

\section{Qualitative Examples}
\Cref{fig:example1} shows an example of Java-to-Python translation by PLBART. The translated code is both syntactically and semantically correct \ie our compiler could successfully parse and build the translated code. It passed 2 test cases out of 10 when executed. The translated code is slightly different from the input Java code. In particular, line 13 in the input Java code is a loop that iterates backward (decreasing order). However, line 12 in the generated python code iterates forward (increasing order). 
If the generated python code were {\tt range(c-1,0,-1)} instead of {\tt range(c-1)}, it would pass all the test cases. 
We attribute such behavior to the fact that {\tt range(*)} is a much more frequent pattern than {\tt range(*,0,-1)} in python code.

\section{TransCoder vs. PLBART}
As we consider TransCoder as the primary baseline of our proposed approach using PLBART, for the sake of fairness, we compare them in terms of model structure and training setting.
\Cref{table:replication_params} presents the comparison. 
TransCoder and PLBART both use the Transformer \cite{vaswani2017attention} architecture, but TransCoder is a twice as large model as PLBART.
Both the models have gone through a two-stage training process. In Stage-1, TransCoder is pre-trained via MLM using 920B tokens, while PLBART is pre-trained via DAE using 87B tokens.
In Stage-2, TransCoder leverages 625M tokens and jointly trained via DAE and BT. In comparison, PLBART is trained via BT using 430M tokens.

\paragraph{Why TransCoder does not suffer from the same language generation issue?}
In Stage-1 pre-training, TransCoder only trains the Transformer Encoder and then initializes a decoder with Encoders' parameters, and the cross attention sub-layers are randomly initialized. We speculate that such random initialization leaves TransCoder unbiased towards generating in the same language as input.
Moreover, PLBART uses language ID token as a prefix to generate in the target language. We noticed that PLBART's decoder disregards the prefix token if not fine-tuned to generate in the target language. On the other hand, TransCoder uses language embeddings with each token in the input. Intuitively, this does not allow the TransCoder's decoder to ignore the language information. For example, with position index ``0'' and language ID ``Python'', TransCoder is more likely to generate ``def'' token and less likely to generate ``static'' or ``int'' since they do not appear in the Python language.
In essence, unlike PLBART, TransCoder does not suffer from the issue of sequence-to-sequence models being unable to generate across languages.

\end{document}